\documentclass[acmsmall]{acmart}

\AtBeginDocument{%
	\providecommand\BibTeX{{%
			\normalfont B\kern-0.5em{\scshape i\kern-0.25em b}\kern-0.8em\TeX}}}

\usepackage{graphicx}

\usepackage[utf8]{inputenc}
\usepackage{amsmath,amssymb,amsfonts}
\usepackage{bbm}
\usepackage{booktabs}
\usepackage{enumitem}
\usepackage{blindtext}
\usepackage{csquotes}
\usepackage{siunitx}
\usepackage{soul}
\usepackage[titletoc,title]{appendix}
\DeclareMathOperator*{\argmax}{arg\,max}

\usepackage{array}
\newcolumntype{L}[1]{>{\raggedright\let\newline\\\arraybackslash\hspace{0pt}}m{#1}}
\newcolumntype{C}[1]{>{\centering\let\newline\\\arraybackslash\hspace{0pt}}m{#1}}
\newcolumntype{R}[1]{>{\raggedleft\let\newline\\\arraybackslash\hspace{0pt}}m{#1}}
\newcommand{\website}{\url{https://www.ipd.kit.edu/ocal}}
\newcommand{\numexperiments}{84,000\xspace}
\usepackage{amsthm}

\theoremstyle{plain}
\newtheorem{example}{Example}
\newtheorem{definition}{Definition}

\setcopyright{none}
\renewcommand\footnotetextcopyrightpermission[1]{}
\begin{document}
	
\title[Overview on One-Class Active Learning for Outlier Detection]{An Overview and a Benchmark of Active Learning for Outlier Detection with One-Class Classifiers}

\author{Holger Trittenbach}
\email{holger.trittenbach@kit.edu}

\author{Adrian Englhardt}
\email{adrian.englhardt@kit.edu}

\author{Klemens Böhm}
\email{klemens.boehm@kit.edu}

\affiliation{%
	\institution{Karlsruhe Institute of Technology (KIT)}
	\streetaddress{Kaiserstr. 12}
	\city{Karlsruhe}
	\postcode{76131}
	\country{Germany}
}

\renewcommand{\shortauthors}{H. Trittenbach, A. Englhardt and K. Böhm}

\begin{abstract}
Active learning methods increase classification quality by means of user feedback.
An important subcategory is active learning for outlier detection with one-class classifiers.
While various methods in this category exist, selecting one for a given application scenario is difficult.
This is because existing methods rely on different assumptions, have different objectives, and often are tailored to a specific use case.
All this calls for a comprehensive comparison, the topic of this article.

This article starts with a categorization of the various methods.
We then propose ways to evaluate active learning results.
Next, we run extensive experiments to compare existing methods, for a broad variety of scenarios.
Based on our results, we formulate guidelines on how to select active learning methods for outlier detection with one-class classifiers.

\end{abstract}


\maketitle

\thispagestyle{empty}

\section{Introduction}

Active learning involves users in machine learning tasks by asking for ancillary information, such as class labels.
Naturally, providing such information requires time and intellectual effort of the users.
To allocate these resources efficiently, active learning employs \emph{query selection} to identify observations for feedback that are likely to benefit classifier training.
In this article, we focus on active learning for outlier detection where so-called one-class classifiers learn to discern between objects from a majority class and unusual observations.
Examples are network security~\cite{Gornitz2009-sa,Stokes2008-gn} or fault monitoring~\cite{Yin2018-jn} where unusual observations like breaches or catastrophic failures are rare to non-existent.

The imbalance between majority-class observations and outliers has important implications on active learning. 
Well-established concepts for query selection, like the margin between two classes, are no longer applicable.
This has motivated specific research on one-class active learning~\cite{Barnabe-Lortie2015-fm,Ghasemi2011-tv,Ghasemi2011-fl,Juszczak2003-ix,Gornitz2013-ed}.
However, as we will show, query selection methods proposed for one-class classifiers differ in their objectives and in the assumptions behind them, and not all of them are suited for outlier detection.
For instance, outliers do not follow a joint distribution, i.e., different outliers may be from different classes.
So active learning methods that rely on density estimation for the minority class are inadequate.
This distinguishes outlier detection from other applications of one-class classification, like collaborative filtering~\cite{Pan2008-ne}.

\begin{figure}[t!]
	\centering
	\includegraphics[width=0.55\linewidth]{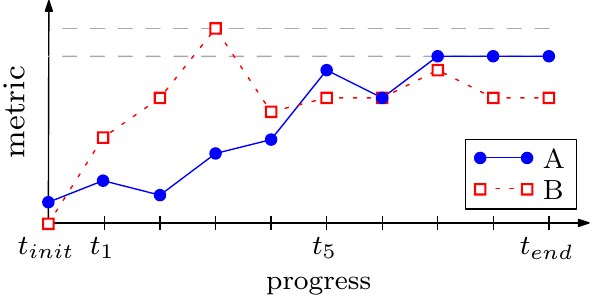}
	\caption{Illustration of an active learning progress curve for active learning methods A and~B.}\label{fig:al_curve_illustrations}
\end{figure}

In addition, evaluation of active learning may lack reliability and comparability~\cite{kottke2017challenges}, in particular with one-class classification.
Evaluations often are use-case specific, and there is no standard way to report results.
This makes it difficult to identify a learning method suitable for a certain use case, and to assess novel contributions in this field.
-- These observations give way to the following questions, which we study in this article:

\begin{enumerate}[leftmargin=8\parindent, listparindent = -\leftmargin , labelsep = 1em, itemindent = 0pt, rightmargin = 2em, topsep = 1ex, itemsep = 0.9ex, labelwidth=6em]
	\item[\emph{Categorization}] What may be a good categorization of learning objectives and assumptions behind one-class active learning?
	\item[\emph{Evaluation}] How to evaluate one-class active learning, in a standardized way?
	\item[\emph{Comparison}] Which active learning methods perform well with outlier detection?
\end{enumerate}

\subsection{Challenges}
Answering these questions is difficult for two reasons.
First, we are not aware of any existing categorization of learning objectives and assumptions.
To illustrate, a typical learning objective is to improve the accuracy of the classifier.
Another, different learning objective is to present a high share of observations from the minority class to the user for feedback~\cite{Das2016-mb}.
In general, active learning methods may perform differently with different learning objectives.
Next, assumptions limit the applicability of active learning methods.
For instance, a common assumption is that some labeled observations are already available before active learning starts.
Naturally, methods that rely on this assumption are only applicable if such labels indeed exist.
So knowing the range of objectives and assumptions is crucial to assess one-class active learning.
Related work however tends to omit respective specifications. 
We deem this one reason why no overview article or categorization is available so far that could serve as a reference point.

Second, there is no standard to report active learning results.
The reason is that \enquote{quality} can have several meanings with active learning, as we now explain.

\begin{example}\label{ex:progress-curve}
Figure~\ref{fig:al_curve_illustrations} is a progress curve.
Such curves are often used to compare active learning methods.
The y-axis is the values of a metric for classification quality, such as the true-positive rate.
The x-axis is the progress of active learning, such as the percentage of observations for which the user has provided a label.
Figure~\ref{fig:al_curve_illustrations} plots two active learning methods A and B from an initial state $t_{init}$ to the final iteration $t_{end}$.
Both methods apparently have different strengths.
A yields better quality at $t_{init}$, while B improves faster in the first few iterations.
However, quality increases non-monotonically, because feedback can bias the classifier temporarily.
At $t_{end}$, the quality of B is lower than the one of~A.
\end{example}

\noindent
The question that follows is which active learning method one should prefer.
One might choose the one with higher quality at $t_{end}$. 
However, the choice of $t_{end}$ is arbitrary, and one can think of alternative criteria such as the stability of the learning rate.
These missing evaluation standards are in the way of establishing comprehensive benchmarks that go beyond comparing individual progress curves.\\

\subsection{Contributions}
This article contains two parts: an overview on one-class active learning for outlier detection, and a comprehensive benchmark of state-of-the-art methods.
We make the following specific contributions.

(i) We propose a categorization of one-class active learning methods by introducing learning scenarios. 
A learning scenario is a combination of a learning objective and an initial setup.
One important insight from this categorization is that the learning scenario and the learning objective are decisive for the applicability of active learning methods.
In particular, some active learning methods and learning scenarios are incompatible.
This suggests that a rigorous specification of the learning scenario is important to assess novel contributions in this field.
We then (ii) introduce several complementary ways to summarize progress curves, to facilitate a standard evaluation of active learning in benchmarks.
The evaluation by progress-curve summaries has turned out to be very useful, since they ease the comparison of active-learning methods significantly.
As such, the categorization and evaluation standards proposed give way to a more reliable and comparable evaluation.

In the second part of our article, we (iii) put together a comprehensive benchmark with around \numexperiments combinations of learning scenarios, classifiers, and query strategies for the selection of one-class active learning methods.
To facilitate reproducibility, we make our implementations, raw results and notebooks publicly available.\footnote{\website} 
A key observation from our benchmark is that none of the state-of-the-art methods stands out in a competitive evaluation.
We have found that the performance largely depends on the parametrization of the classifier, the data set, and on how progress curves are summarized.
In particular, a good parametrization of the classifier is as important as choosing a good query selection strategy.
We conclude by (iv) proposing guidelines on how to select active learning methods for outlier detection with one-class classifiers.

\section{Overview on One-Class Active Learning}\label{sec:ocal}

One-class classification is a machine learning method that is popular in different domains.
Thus, we fix some terminology before we review the concepts of one-class active learning.
We then address Question \emph{Categorization} with a discussion of the building blocks and assumptions of one-class active learning.

\subsection{Terminology}
In this article, we focus on \emph{one-class classification for outlier detection}.
This is a subset of the broader class \emph{one-class classification}, which includes other applications, like collaborative filtering~\cite{Pan2008-ne}.
The objective of one-class classification for outlier detection is to learn a decision function that discerns between normal and unusual observations.
What constitutes a normal and an unusual class may depend on the context, see \autoref{sec:specific-assumptions}.

One may additionally distinguish between categories of one-class classifiers.
One category is unsupervised one-class classifiers, which learn a decision without any class label information.
If one-class classifiers make use of such information, they fall into the category of semi-supervised learning.
A special case of semi-supervised methods is learning from positive and unlabeled observations~\cite{Li2005-sn}.

There are different ways to design one-class active learning (AL) systems, and several variants have recently been proposed.
Yet we have found that variants follow different objectives and make implicit assumptions.
Existing surveys on active learning do not discuss these objectives and assumptions, and they rather focus on general classification tasks~\cite{Ramirez-Loaiza2017-iq,settles2012active,beyer2015select,Olsson2009-hd} and on benchmarks for balanced~\cite{bernard2018towards} and multi-class classification~\cite{Juszczak2003-ix}.

\begin{figure}[t!]
	\centering
	\includegraphics[width=0.7\linewidth]{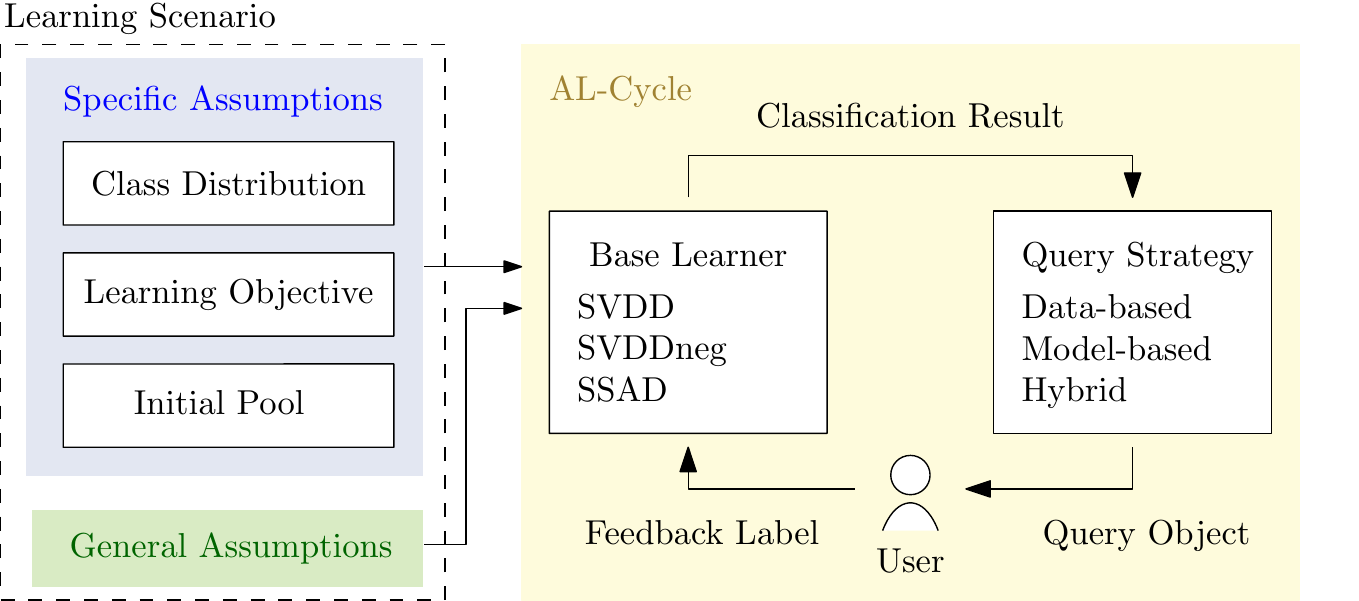}
	\caption{Building blocks of one-class active learning.}\label{fig:ocal_overview}
\end{figure}

In the remainder of this section, we discuss assumptions for one-class AL, structure the aspects where one-class AL systems differ from each other, and discuss implications of design choices on the AL system.
We structure our discussion into three parts corresponding to the building blocks of a one-class AL system.
Figure~\ref{fig:ocal_overview} graphs the building blocks.
The first block is the \emph{AL Setup}, which establishes assumptions regarding the training data and the process of gathering user feedback.
It specifies the initial configuration of the system before the actual active learning starts.
The second building block is the \emph{Base Learner}, i.e., a one-class classifier that learns a binary decision function based on the data and user feedback available.
The third building block is the \emph{Query Strategy}.
It is a method to select observations that a user is asked to provide feedback for.

We call observations that a query strategy selects \emph{query objects}, the entity that provides the label an \emph{oracle}, and the process of providing label information \emph{feedback}.
In a real scenario, the oracle is a user.
For benchmarks, the oracle is simulated, based on a given gold standard.
In what follows, we explain the blocks and discuss dependencies between them.

\subsection{Building Block: Learning Scenario}
\label{sec:learning-scenario}

Researchers make assumptions regarding the interaction between system and user as well as assumptions regarding the application scenario.
Literature on one-class AL often omits an explicit description of these assumptions, and one must instead derive them for instance from the experimental evaluation.
Moreover, assumptions often do not come with an explicit motivation, and the alternatives are unclear.

We now review the various assumptions found in the literature.
We distinguish between two types, general and specific assumptions.

\subsubsection{General Assumptions}
General assumptions specify modalities of the feedback and impose limits on how applicable AL is in real settings.
These assumptions have been discussed for standard binary classification \cite{settles2012active}, and many of them are accepted in the literature.
We highlight the ones important for one-class~AL.

\emph{Feedback Type:}
Existing one-class AL methods assume that feedback is a class label, i.e., the decision whether an observation belongs to a class or not.
However, other types of feedback are conceivable as well, such as feature importance \cite{Raghavan2006-ps,Druck2009-ep}.
But to our knowledge, research on one-class AL has been limited to label feedback.
Next, the most common mechanism in literature is sequential feedback, i.e., for one observation at a time.
However, asking for feedback in batches might have certain advantages, such as increased efficiency of the labeling process.
But a shift from sequential to batch queries is not trivial and requires additional diversity criteria \cite{Juszczak2006-ut}.

\emph{Feedback Budget:}
A primal motivation for active learning is that the amount of feedback a user can provide is bounded.
For instance, the user can have a time or cost budget or a limited attention span to interact with the system.
Assigning costs to feedback acquisition is difficult, and a budget restriction is likely to be application-specific.
In some cases, feedback on observations from the minority class may be more costly.
However, a common simplification here is to assume that labeling costs are uniform, and that there is a limit on the number of feedback iterations.

\emph{Interpretability:}
A user is expected to have sufficient domain knowledge to provide feedback purposefully.
However, this implies that the user can interpret the classification result in the first place, i.e., the user understands the output of the one-class classifier.
This is a strong assumption, and it is difficult to evaluate.
For one thing, \enquote{interpretation} already has various meanings for non-interactive supervised learning~\cite{Lipton2016-wa}, and it has only recently been studied for interactive learning \cite{Phillips2018-ys,Teso2018-jo}.
Concepts to support users with an explanation of outliers \cite{Micenkova2013-hr,Kauffmann2018-dr} have not been studied in the context of active learning either.
In any case, a thorough evaluation would require a user study.
However, existing one-class AL systems bypass the difficulty of interpretation and assume a perfect oracle, i.e., an oracle which provides feedback with a predefined accuracy.\\

\subsubsection{Specific Assumptions}\label{sec:specific-assumptions}
Specific assumptions confine the learning objective and the data for a particular AL application.
One must define specific assumptions carefully, because they restrict which base learners and query strategies are applicable.
We partition specific assumptions into the following categories.

\emph{Class Distribution:}
One-class learning is designed for high\-ly imbalanced domains.
There are two different definitions of \enquote{minority class}.
The first one is that the minority class is unusual observations, also called outliers, that are exceptional in a bulk of data.
The second definition is that the minority class is the target in a one-vs-all multi-class classification task, i.e., where all classes except for the minority class have been grouped together \cite{Juszczak2003-ko,Ghasemi2011-fl}.
With this definition, the minority class is not exceptional, and it has a well-defined distribution.
Put differently, one-class classification is an alternative to imbalanced binary classification in this case.
So both definitions of \enquote{minority class} relate to different problem domains.
The first one is in line with the intent of our paper, and we stick to it in the following.

Under the first definition, one can differentiate between characterizations of outliers.
The prevalent characterization is that outliers do not follow a common underlying distribution.
This assumption has far-reaching implications.
For instance, if there is no joint distribution, it is not meaningful to estimate a probability density from a sample of the minority class.

Another characterization of outliers is to assume that it is a mixture of several distributions of rare classes.
In this case, a probability density for each mixture component exists.
So the probability density for the mixture as a whole exists as well.
Its estimation however is hard, because the sample for each component is tiny.
The characterization of the outlier distribution has implications on the separation of the data into train and test partitions, as we will explain in~\autoref{sec:split-strategy}.

\emph{Learning Objective:}
The learning objective is the benefit expected from an AL system.
A common objective is to improve the accuracy of a classifier.
But there are alternatives.
For instance, users of one-class classification often have a specific interest in the minority class~\cite{Das2016-mb}.
In this case, it is reasonable to assume that users prefer giving feedback on minority observations if they will examine them anyhow later on. 
So a good active learning method yields a high proportion of queries from the minority class.
This may contradict the objective of accuracy improvement.

There also are cases where the overall number of available observations is small, even for the majority class.
The learning objective in this case can be a more robust estimate of the majority-class distribution~\cite{Ghasemi2011-fl,Ghasemi2011-tv}.
A classifier benefits from extending the number of majority-class labels.
This learning objective favors active learning methods that select observations from the majority class.

\emph{Initial Pool:}
The initial setup is the label information available at the beginning of the AL process.
There are two cases: (i)~Active learning starts from scratch, i.e., there are no labeled examples, and the initial learning step is unsupervised. (ii)~There are some labeled instances available~\cite{Juszczak2006-ut}.
The number of observations and the share of class labels in the initial sample depends on the sampling mechanism.
A special case is if the labeled observations exclusively are from the majority class~\cite{Ghasemi2011-fl}.
In our article, we consider different initial pool strategies:
\begin{enumerate}[leftmargin =3.8\parindent, rightmargin = 2em, topsep = 1ex, itemsep = 0.5ex, align=left, labelwidth=2.3em]
	\item[(Pu)] \underline{P}ool \underline{u}nlabeled: All observations are unlabeled.
	\item[(Pp)] \underline{P}ool \underline{p}ercentage: Stratified proportion of labels for $p$ percent of the observations.
	\item[(Pn)] \underline{P}ool \underline{n}umber: Stratified proportion of labels for a fixed number of observations $n$.
	\item[(Pa)] \underline{P}ool \underline{a}ttributes: As many labeled inliers as number of attributes.
\end{enumerate}
The rationale behind Pa is that the correlation matrix of labeled observations is singular if there are fewer labeled observations than attributes.
With a singular correlation matrix, some query strategies are infeasible.\\

How general and specific assumptions manifest depends on the use case, and different combinations of assumptions are conceivable.
We discuss how we set assumptions for our benchmark in~\autoref{sec:experiments-and-results}.

\subsection{Notation}
\newcommand{\inlier}{\text{in}\xspace}
\newcommand{\outlier}{\text{out}\xspace}

Before we introduce the remaining two building blocks, we specify some notation.
$\mathcal{X}\subseteq\mathbb{R}^M$ is a data space with $M$ attributes.
$\mathbf{X}$ is a sample from $\mathcal{X}$ of $n$ observations $\{x_1, x_2, ..., x_n\}$, where each observation $x_i$ is a vector of $m$~attribute values, i.e., $x_i = (x_{i1}, x_{i2}, \dots, x_{im})$.
In this article, each observation either belongs to the minority or to the majority class.
For brevity, we call an observation from the minority class \emph{outlier} and one from the majority class \emph{inlier}, and we encode them with a categorical class label $y_i\in\{\inlier, \outlier\}$.
Synonyms for inlier are \enquote{target}, \enquote{positive observation} or \enquote{regular observation}, and for outlier \enquote{anomalous observation} or \enquote{exceptional observation}.
$\mathbf{X}$ can be partitioned into the unlabeled set of observations $\mathcal{U}$, i.e., observations where $y_i$ is unknown, and the labeled set of observations $\mathcal{L}$.
We distinguish between the labeled inliers $\mathcal{L_\inlier}$ and the labeled outliers $\mathcal{L_\outlier}$.

\subsection{Building Block: Base Learner}
\label{sec:base-learner}

A base learner is a one-class classifier that discerns between inliers and outliers.
It takes observations and a set of class labels as input and returns a decision function.

\begin{definition}[Decision Function]
	\label{def:base-learner-decision-function}
	A decision function $f$ is a function of type $f\colon \mathcal{X} \rightarrow \mathbb{R}$.
	An observation is assigned to the minority class if $f(x) > 0$ and to the majority class otherwise.
\end{definition}
\noindent
One-class classifiers fall into two categories: support-vector methods and non-support-vector classifiers \cite{Khan2014-lo}.
In our article, we focus on support-vector methods, the prevalent choice as base learners for one-class AL.
However, the query strategy is independent from a specific instantiation, as long as the base learner returns a decision function.

One can further distinguish between semi-supervised and unsupervised one-class classifiers.
Both have been used with one-class AL, but whether they are applicable depends on the learning scenario.
A \emph{semi-supervised} base learner uses both unlabeled data and labeled data with class labels for training.
Labels can either come from both classes or only from the minority class~\cite{Tax2004-ss}.
An \emph{unsupervised} base learner does not have any mechanism to use label information directly to train the decision function.
Instead, one can manipulate the unsupervised base learner by exposing it only to specific subsets of the training data. 
For instance, one can train on labeled inliers only.

In this current article, we restrict our discussion to base learners that have been used in previous work on active learning for outlier detection with one-class classifiers.
In particular, we use unsupervised SVDD~\cite{Tax2004-ss}, semi-supervised SVDDneg~\cite{Tax2004-ss} with labels from the minority class, and the semi-supervised SSAD~\cite{Gornitz2013-ed} with labels from both classes.

\subsubsection{Support Vector Data Description (SVDD)}

One of the most popular support-vector methods is Support Vector Data Description (SVDD)~\cite{Tax2004-ss}.
The core idea is to fit a sphere around the data that encompasses all or most observations.
This can be expressed as a Minimum Enclosing Ball (MEB) optimization problem of the following form
\begin{equation}\label{eq:svdd-primal}
\begin{aligned}
& \underset{a,R,\mathbb{\xi}}{\text{minimize}} & & R^2 + C \sum_{i=1}^{N} \xi_i \\
& \text{subject to}	& & \lVert \phi(x_i) - a \rVert^2 \leq R^2 + \xi_i, \; i = 1, \ldots, N. \\
& & & \xi_i \geq 0, \; i = 1, \ldots, N.
\end{aligned}
\end{equation}
with the center of the ball $a$, the radius $R$, slack variables $\xi_i$, a cost parameter $C \in (0,1]$, and a function $\phi\colon\mathcal{X} \rightarrow \mathcal{F}$ which maps $x$ into a reproducing kernel Hilbert space $\mathcal{F}$.
Solving the optimization problem gives a fixed $a$ and $R$ and the decision function
\begin{equation}\label{eq:decision-function}
f(x_i) = \lVert \phi(x_i) - a \rVert - R .
\end{equation}
The slack variables $\xi_i$ relax the MEB, i.e., they introduce a trade-off to allow observations to fall outside the sphere at cost $C$.
If $C$ is high, observations falling outside of the sphere are expensive.
In other words, $C$ controls the share of objects that are outside the decision boundary.

The optimization problem from \autoref{eq:svdd-primal} can be solved in the dual space.
In this case, the problem contains only inner products of the form $\langle \phi(x), \phi(x') \rangle$.
This allows to use the kernel trick, i.e., to replace the inner products with a kernel function $k(x,x') \rightarrow \mathbb{R}, x, x' \in \mathcal{X}$.
A common kernel is the Radial Basis Function (RBF) Kernel
\begin{equation}
k_{\text{RBF}}(x,x') = e^{-\gamma \lVert x - x'\rVert^2}.
\end{equation}
with parameter $\gamma$.
Larger $\gamma$ values correspond to more flexible decision boundaries.

\subsubsection{SVDD with Negative Examples (SVDDneg)}

SVDDneg \cite{Tax2004-ss} extends the vanilla SVDD by using different costs $C_1$ for $\mathcal{L}_\inlier$ and $\mathcal{U}$ and costs $C_2$ for $\mathcal{L}_\outlier$.
An additional constraint places observations in $\mathcal{L_{\outlier}}$ outside the decision boundary.
\begin{equation}\label{eq:svddneg-primal}
\begin{aligned}
& \underset{a,R,\mathbb{\xi}}{\text{minimize}} & & R^2 + C_1 \cdot \sum_{i\colon x_i \in \mathcal{U} \cup \mathcal{L_\inlier}} \xi_i +  C_2 \cdot \sum_{i\colon x_i \in \mathcal{L_\outlier}} \xi_i\\
& \text{subject to}	& & \lVert \phi(x_i) - a \rVert^2 \leq R^2 + \xi_i, \; i\colon x_i \in \mathcal{U} \cup \mathcal{L_\inlier} \\
& & & \lVert \phi(x_i) - a \rVert^2 \geq R^2 - \xi_i, \; i\colon x_i \in \mathcal{L_\outlier} \\
& & & \xi_i \geq 0, \; i = 1, \ldots, N.
\end{aligned}
\end{equation}

\subsubsection{Semi-Supervised Anomaly Detection (SSAD)}
SSAD~\cite{Gornitz2013-ed} additionally differentiates between labeled inliers and unlabeled observations in the objective and in the constraints.
In its original version, SSAD assigns different costs to $\mathcal{U}$, $\mathcal{L}_\inlier$, and $\mathcal{L}_\outlier$.
We use a simplified version where the cost for both $\mathcal{L}_\inlier$ and $\mathcal{L}_\outlier$ are $C_2$.
SSAD further introduces an additional trade-off parameter, which we call~$\kappa$.
High values of $\kappa$ increase the weight of $\mathcal{L}$ on the solution, i.e., SSAD is more likely to overfit to instances in~$\mathcal{L}$.

\begin{equation}\label{eq:sadd-primal}
\begin{aligned}
& \underset{a,R,\mathbb{\xi}, \tau}{\text{minimize}} & & R^2 - \kappa \tau + C_1 \cdot \sum_{i\colon x_i \in \mathcal{U}} \xi_i + C_2 \cdot \sum_{i\colon x_i \in \mathcal{L}} \xi_i\\
& \text{subject to}	& & \lVert \phi(x_i) - a \rVert^2 \leq R^2 + \xi_i, \; i\colon x_i \in \mathcal{U} \\
& 	& & \lVert \phi(x_i) - a \rVert^2 \geq R^2 - \xi_i + \tau , \; i\colon x_i \in \mathcal{L_{\outlier}} \\
& 	& & \lVert \phi(x_i) - a \rVert^2 \leq R^2 + \xi_i - \tau, \; i\colon x_i \in \mathcal{L_{\inlier}} \\
& & & \xi_i \geq 0, \; i = 1, \ldots, N.
\end{aligned}
\end{equation}
Under mild assumptions, SSAD can be reformulated as a convex problem \cite{Gornitz2013-ed}.\\

Parameterizing the kernel function and cost parameters is difficult, because a good parameterization typically depends on the data characteristics and the application.
Further, one has to rely on heuristic to find a good parametrization in unsupervised settings.
There are several heuristics to find a good parametrization that use data characteristics~\cite{silverman2018density,scott2015multivariate,xiao2014two}, artificial outliers~\cite{tax2001uniform,banhalmi2007counter,wang2018hyperparameter}, or SVDD-specific properties like the number of support vectors~\cite{wang2013modified}.
However, optimizing the parametrization is not a focus of this article, and we rely on established methods to select the kernel and the cost parameters, see \autoref{app:experimental_setup}.

\subsection{Building Block: Query Strategy}\label{sec:query-strategies}

A query strategy is a method that selects observations for feedback.
In this section, we review the respective principles, as well as existing strategies for one-class classification.
The varying notation in the literature would make an overview difficult to follow.
So we rely on notation introduced earlier.

To decide on observations for feedback, query strategies rank unlabeled observations according to an informativeness measure.
\begin{definition}[Informativeness]
	Let a decision function $f$, unlabeled observations $\mathcal{U}$ and labeled observations $\mathcal{L}$ be given.
	Informativeness is a function $x \mapsto \tau(x, \mathcal{U},\mathcal{L}, f)$ that maps an observation $x \in \mathcal{U}$ to $\mathbb{R}$.
\end{definition}
\noindent
For brevity, we only write $\tau(x)$.
$\tau(x)$ quantifies how valuable feedback for observation $x \in \mathcal{U}$ is for the classification model.
This definition is general, and there are different ways to interpret \emph{valuable}.
Feedback can be valuable if the model is uncertain with the prediction of an observation, or if the classification error is expected to decrease.
Some query strategies also balance between the representativeness of observations and the exploration of sparse regions.
In this case, local density estimates affect the value of an observation.

In general, a query strategy selects one or more observations based on their informativeness.
We define it as follows.
\begin{definition}[Query Strategy]
	A query strategy $\mathit{QS}$ is a function of type \[\mathit{QS}\colon \mathcal{U} \times \mathbb{R} \rightarrow \mathcal{Q}\] with $\mathcal{Q} \subseteq \mathcal{U}$. 
\end{definition}
\noindent
The feedback on $\mathcal{Q}$ from an oracle results in an updated set of labeled $\mathcal{L'} = \mathcal{L} \cup \mathcal{Q}$ and unlabeled data $ \mathcal{U'} = \mathcal{U} \setminus \mathcal{Q}$.
In this current article, we only consider single queries (cf.\ \autoref{sec:learning-scenario}).
Given this, we assume query strategies to always return the observation with the highest informativeness
\begin{equation}
\label{eq:selection}
\mathcal{Q}	= \argmax_{x \in U} \tau(x).
\end{equation}

We now review existing query strategies from literature that have been proposed for one-class active learning.
To this end, we partition them into three categories.
The first category is \emph{data-based query strategies}.\footnote{Others have used the term \enquote{model-free} instead~\cite{ONeill2017-ng}. 
However, we deliberately deviate from this nomenclature since the strategies we discuss still rely on some kind of underlying model, e.g., a kernel-density estimator.}
These strategies approach query selection from a statistical side.
The second category is \emph{model-based query strategies}.
These strategies rely on the decision function returned by the base learner.
The third category is \emph{hybrid query strategies}.
These strategies use both the data statistics and the decision function.\\

\subsubsection{Data-based Query Strategies}
The concept behind data-based query strategies is to compare the posterior probabilities of an observation $p(\outlier | x)$ and $p(\inlier | x)$.
This is well known from binary classification and is referred to as \emph{measure of uncertainty} \cite{settles2012active}.
If a classifier does not explicitly return posterior probabilities, one can use the Bayes rule to infer them.
But this is difficult, for two reasons.
First, applying the Bayes rule requires knowing the prior probabilities for each class, i.e., the proportion of outliers in the data.
It may not be known in advance.
Second, outliers do not follow a homogeneous distribution. 
This renders estimating $p(x|\outlier)$ infeasible.
There are two types of data-based strategies that have been proposed to address these difficulties.

The first type deems observations informative if the classifier is uncertain about their class label, i.e., observations with equal probability of being classified as inlier and outlier.
The following two strategies quantify informativeness in this way.\\

\emph{Minimum Margin} \cite{Ghasemi2011-tv}: This QS relies on the difference between posterior class probabilities
\begin{subequations}
\begin{align}
\tau_{\text{MM}}(x) &= - |p(\inlier|x) - p(\outlier|x)| \label{eq:tau_MM_1} \\
&= -\left|\frac{p(x|\inlier) \cdot p(\inlier) - p(x|\outlier) \cdot p(\outlier)}{p(x)}\right| \label{eq:tau_MM_2}\\
&= -\left|\frac{2 \cdot p(x|\inlier) \cdot p(\inlier) - p(x)}{p(x)}\right|. \label{eq:tau_MM_3}
\end{align}
\end{subequations}
where \autoref{eq:tau_MM_2} and \autoref{eq:tau_MM_3} follow from the Bayes rule.
If $p(\text{\inlier})$ and $p(\text{\outlier})$ are known priors, one can make direct use of \autoref{eq:tau_MM_3}.
Otherwise, the inventors of Minimum Margin suggest to take the expected value under the assumption that $p(\text{\outlier})$, i.e., the share of outliers, is uniformly distributed
\begin{equation}\label{eq:tau_EMM}
\tau_{\text{EMM}}(x) = \mathbb{E}_{p(\outlier)}\left( \tau_{\text{MM}}(x) \right) = \left(\frac{p(x|\inlier)}{p(x)} -1\right) \cdot sign \left( 0.5 - \frac{p(x|\inlier)}{p(x)} \right).
\end{equation}
We find this an unrealistic assumption, because a share of outliers of 0.1 would be as likely as 0.9.
In our experiments, we evaluate both $\tau_{\text{MM}}$ with the true outlier share as a prior and with $\tau_{\text{EMM}}$.\\

\emph{Maximum-Entropy} \cite{Ghasemi2011-tv}: This QS selects observations where the distribution of the class probability has a high entropy
\begin{equation}\label{eq:tau_ME}
\tau_{\text{ME}}(x) = -[ p(\inlier|x) \cdot \log(p(\inlier|x)) + p(\outlier|x) \cdot \log(p(\outlier|x)) ].
\end{equation}
Applying the Bayes rule and taking the expected value as in \autoref{eq:tau_EMM} gives
\newcommand{\ax}{\frac{p(x|\inlier)}{p(x)}}
\begin{equation}\label{eq:tau_EME}
\begin{aligned}
\tau_{\text{EME}}(x) &= \mathbb{E}_{p(\outlier)}\left( \tau_{\text{ME}}(x) \right) \\
&= \frac{- \left(\ax \right)^2 \cdot \log\left(\ax\right) \ + \ax}{2\cdot \ax} \\
& + \frac{\left(\ax-1\right)^2 \cdot \log\left(1-\ax\right)}{2\cdot \ax}.
\end{aligned}
\end{equation}

\begin{figure}
	\centering
		\includegraphics[width=0.65\linewidth]{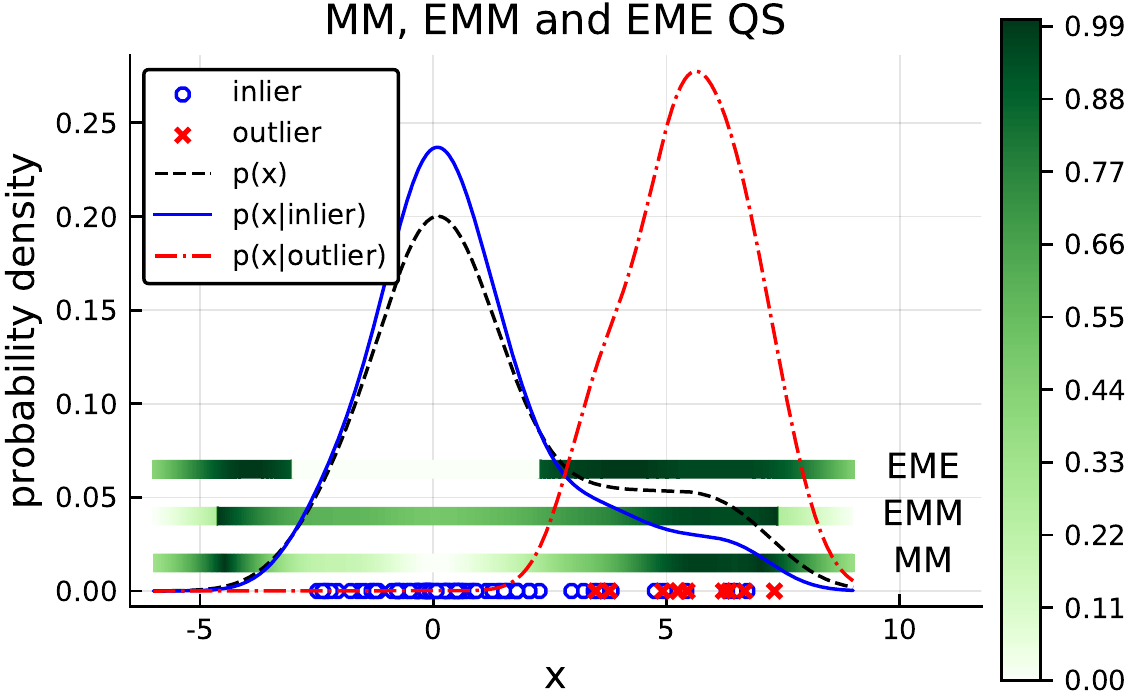}
		\caption{Visualization of informativeness calculated by $\tau_{\text{MM}}$ (\autoref{eq:tau_MM_3}), $\tau_{\text{EMM}}$ (\autoref{eq:tau_EMM}) and $\tau_{\text{EME}}$ (\autoref{eq:tau_EME}).
			Dark colored regions indicate regions of high informativeness.}
		\label{fig:plot_ghasemi_probability_densities}
\end{figure}

To give an intuition of the Minimum-Margin and the Maxi\-mum-Entropy strategy, we visualize the informativeness for Minimum Margin and Maximum Entropy on sample data.
\autoref{fig:plot_ghasemi_probability_densities} visualizes $\tau_{\text{MM}}$, $\tau_{\text{EMM}}$ and $\tau_{\text{ME}}$ for univariate data generated from two Gaussian distributions, with $p(\outlier) = 0.1$.
The authors of $\tau_{\text{EME}}$ suggest to estimate the densities with kernel density estimation (KDE) \cite{Ghasemi2011-tv}.
However, entropy is defined on probabilities and is not applicable to densities, so just inserting into the formula yields ill-defined results.
Moreover, $\tau_{\text{EME}}$ is not defined for $\ax \geq 1$.
We set $\tau_{\text{EME}} = 0$ in this case.
For $\tau_{\text{MM}}$, we use \autoref{eq:tau_MM_3} with prior class probabilities.
Not surprisingly, all three depicted formulas result in a similar pattern, as they follow the same general motivation.
The tails of the inlier distribution yield high informativeness.
The informativeness decreases slower on the right tail of the inlier distribution where the outlier distribution has some support.

The second type of data-based query strategies strives for a robust estimation of the inlier density.
The idea is to give high informativeness to observations that are likely to reduce the loss between the estimated and the true inlier density.
There is one strategy of this type.

\emph{Minimum-Loss} \cite{Ghasemi2011-fl}: Under the minimum-loss strategy, observations have high informativeness if they are expected to increase the estimate of the inlier density.
The idea to calculate this expected value is as follows.
The feedback for an observation is either \enquote{outlier} or \enquote{inlier}.
The minimum-loss strategy calculates an updated density for both cases and then takes the expected value by weighting each case with the prior class probabilities.
Similarly to \autoref{eq:tau_MM_3}, this requires knowledge of the prior class probabilities.

We now describe Minimum-Loss formally.
Let $\hat{g}^{\inlier}$ be an estimated probability density over all inlier observations $\mathcal{L_\inlier}$.
Let $\mathcal{L}^{x}_\inlier = \mathcal{L}_\inlier \cup \{x\}$, and let $\hat{g}^{\inlier, x}$ be its corresponding density. 
Similarly, we define $\mathcal{L}^{x}_\outlier = \mathcal{L}_\outlier \cup \{x\}$.
Then $\hat{g}^{\inlier}_{-i}$ stands for the density estimated over all $\mathcal{L}_\inlier\! \setminus \! x_i$ and $\hat{g}^{\inlier, x}_{-i}$ for $\mathcal{L}^{x}_\inlier\! \setminus \! x_i$ respectively.
In other words, for $\hat{g}^{\inlier, x}_{-i}(x_i)$, one first estimates the density $\hat{g}^{\inlier, x}_{-i}$ without $x_i$ and then evaluates the estimated density at $x_i$.
One can now calculate how well an observation $x$ matches the inlier distribution by using leave-out-one cross validation for both cases.

Case 1: x is inlier
\begin{equation}\label{eq:tau-cv-in}
\tau_{\text{ML-\inlier}}(x) = \frac{1}{\lvert \mathcal{L}^{x}_\inlier \rvert} \sum_{i\colon x_i \in \mathcal{L}^{x}_\inlier} {\hat{g}^{\inlier, x}_{-i}(x_i)} - \frac{1}{\lvert \mathcal{L}_\outlier \rvert}\sum_{i\colon x_i \in \mathcal{L}_\outlier}\hat{g}^{\inlier, x}(x_i).
\end{equation}

Case 2: x is outlier
\begin{equation}\label{eq:tau-cv-out}
\tau_{\text{ML-\outlier}}(x) = \frac{1}{\lvert \mathcal{L}_\inlier \rvert} \sum_{i\colon x_i \in \mathcal{L}_\inlier} {\hat{g}^{\inlier}_{-i}(x_i)} - \frac{1}{\lvert \mathcal{L}^{x}_\outlier \rvert}\sum_{i\colon x_i \in \mathcal{L}^{x}_\outlier}\hat{g}^{\inlier}(x_i).
\end{equation}
\noindent
The expected value over both cases is
\begin{equation}\label{eq:tau-ML}
	\tau_{\text{ML}}(x) = p(\inlier) \cdot \tau_{\text{ML-\inlier}}(x) + (1 - p(\inlier)) \cdot \tau_{\text{ML-\outlier}}(x).
\end{equation}

\begin{figure}
	\centering
	\includegraphics[width=0.65\linewidth]{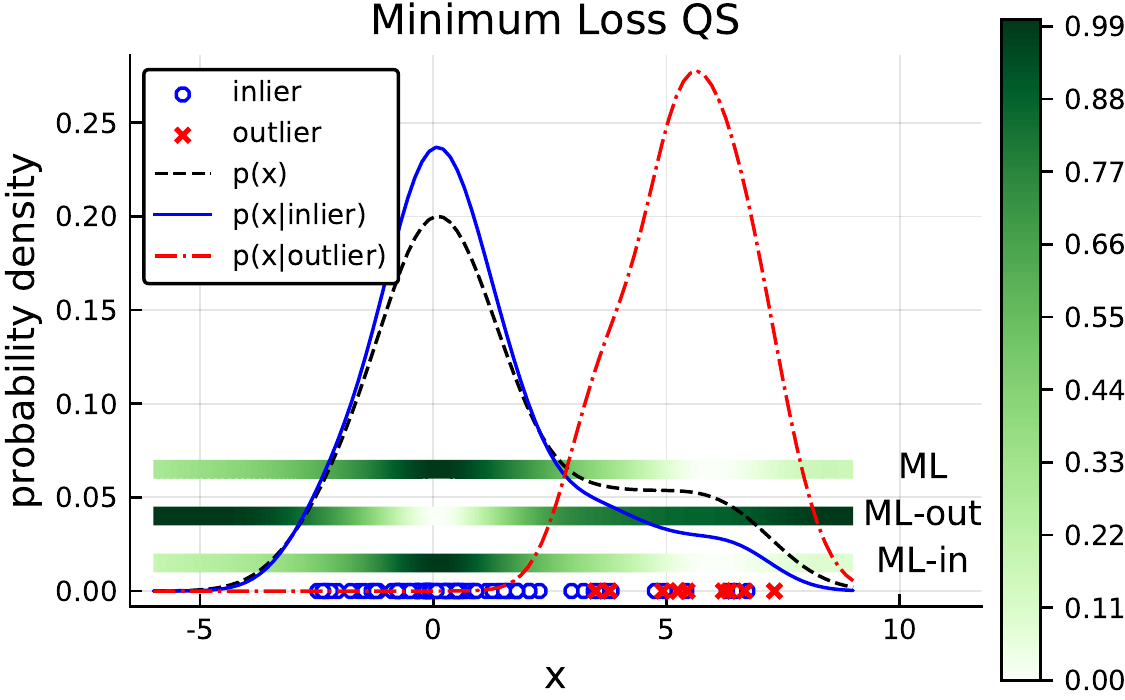}
	\caption{Visualization of informativeness calculated by $\tau_{\text{ML-\inlier}}$ (\autoref{eq:tau-cv-in}), $\tau_{\text{ML-\outlier}}$ \autoref{eq:tau-cv-out}, and $\tau_{\text{ML}}$ \autoref{eq:tau-ML}.
		Dark colored regions indicate regions of high informativeness.}
	\label{fig:plot_minimum_loss_qs}
\end{figure}

\noindent
We illustrate \autoref{eq:tau-cv-in}, \autoref{eq:tau-cv-out} and $\tau_{\text{ML}}$ in \autoref{fig:plot_minimum_loss_qs}.
As expected, $\tau_{\text{ML}}$ yields high informativeness in regions of high inlier density.
$\tau_{\text{ML}}$ gives an almost inverse pattern compared to the Minimum-Margin and the Maximum-Entropy strategies.
This illustrates that existing query strategies are markedly different.
It is unclear how to decide between them solely based on theoretical considerations, and one has to study them empirically instead.\\

\subsubsection{Model-based Query Strategies}
Model-based strategies rely on the decision function $f$ of a base learner.
Recall that an observation $x$ is an outlier if $f(x) > 0$ and an inlier for $f(x) \leq 0$.
Observations with $f(x) = 0$ are on the decision boundary.\\

\emph{High-Confidence} \cite{Barnabe-Lortie2015-fm}: This QS selects observations that match the inlier class the least. 
For SVDD this is
\begin{equation}
	\tau_{\text{HC}}(x) = f(x).
\end{equation}

\emph{Decision-Boundary}: This QS selects observations closest to the decision boundary
\begin{equation}
\tau_{\text{DB}}(x) = - |f(x)|.
\end{equation}

\subsubsection{Hybrid Query Strategies}
Hybrid query strategies combine data-based and model-based strategies.\\

\emph{Neighborhood-Based} \cite{Gornitz2013-ed}: This QS explores unknown neighborhoods in the feature space.
The first part of the query strategy calculates the average number of labeled instances among the k-nearest neighbors
\begin{equation}
\widehat\tau_{\text{NB}}(x) = -\left( 0.5 + \frac{1}{2k} \cdot |\lbrace x' \in \text{NN}_k(x)\colon x' \in \mathcal{L}_{\inlier}\rbrace| \right),
\end{equation}
with k-nearest neighbors $NN_k(\cdot)$.
A high number of neighbors in $\mathcal{L}_{\inlier}$ makes an observation less interesting.
The strategy then combines this number with the distance to the decision boundary, i.e., $\tau_{\text{NB}} = \eta \cdot \tau_{\text{DB}} + (1-\eta) \cdot \widehat\tau_{\text{NB}}$. Parameter $\eta \in \left[0, 1\right]$ controls the influence of the number of already labeled instances in the neighborhood on the decision.
The authors do not recommend any specific parameter value, and we use $\eta = 0.5$ in our experiments.\\

\emph{Boundary-Neighbor-Combination} \cite{Yin2018-jn}: The core of this query strategy is a linear combination of the normalized distance to the hypersphere and the normalized distance to the first-nearest neighbor
\begin{equation}\label{eq:al-strategy:yin}
\begin{split}
\widehat\tau_{\text{BNC}}(x) = & \ (1-\eta) \cdot
\left(
-\frac{ |f(x)| - \min\limits_{x' \in U} {|f(x')|} }
{ \max\limits_{x' \in \mathcal{U}}{|f(x')|}}
\right) \\
& + \eta \cdot \left(
-\frac{ d(x, \text{NN}_1(x)) - \min\limits_{x' \in \mathcal{U}}(d(x', \text{NN}_1(x')) }
{\max\limits_{x' \in \mathcal{U}}d(x', \text{NN}_1(x'))}
\right).
\end{split}
\end{equation}
with a distance function $d$, and trade-off parameter $\eta$.
The actual query strategy $\tau_{\text{BNC}}$ is to choose a random observation with probability p and to use strategy $\widehat\tau_{\text{BNC}}$ with probability $(1-p)$.
The authors recommend to set $\eta = 0.7$ and $p = 0.15$.\\

\subsubsection{Baselines}

In addition to the strategies introduced so far, we use the following baselines.\\

\emph{Random}: This QS draws each unlabeled observation with equal probability
\begin{equation}
\tau_{\text{rand}}(x) = \frac{1}{\lvert \mathcal{U} \rvert} \ .
\end{equation}

\emph{Random-Outlier}: This QS is similar to Random, but with informativeness 0 for observations predicted to be inliers
\begin{equation}
\tau_{\text{rand-out}}(x) = \left\{
\begin{array}{ll} 
\frac{1}{\lvert \mathcal{U} \rvert} &\text{if}\  f(x) > 0 \\
0 &\text{otherwise}.
\end{array} 
\right.
\end{equation}

\noindent
In general, adapting other strategies from standard binary active learning is conceivable as well.
For instance, one could learn a committee of several base learners and use disagreement-based query selection~\cite{settles2012active}.
In this current article however, we focus on strategies that have been explicitly adapted to and used with one-class active learning.

\section{Evaluation of One-Class Active Learning}\label{sec:method-evaluation}

Evaluation of active learning methods is more involved than the one of static methods.
Namely, the result of an AL method is not a single number, but rather a sequence of numbers that result from a quality evaluation in each iteration.

We now address Question \emph{Evaluation} in several steps.
We first discuss characteristics of active learning progress curves.
We then review common quality metrics~(QM) for one-class classification, i.e., metrics that take the class imbalance into account.
We then discuss different ways to summarize active learning curves.
Finally, we discuss the peculiarities of common train/test-split strategies for evaluating one-class active learning and limitations of the design choices just mentioned.

\subsection{Progress Curves}
The sequence of quality evaluations can be visualized as a \emph{progress curve}, see~\autoref{fig:al_curve_illustrations}.
We call the interval from $t_{init}$ to $t_{end}$ an \emph{active learning cycle}.
Literature tends to use the percentage or the absolute number of labeled observations to quantify progress on the x-axis.
However, this percentage may be misleading if the total number of observations varies between data sets.
Next, other measures are conceivable as well, such as the time the user spends to answer a query.
While this might be even more realistic, it is very difficult to validate.
We deem the absolute number of labeled objects during the active learning cycle the most appropriate scaling.
It is easy to interpret, and the budget restriction is straightforward.
However, the evaluation methods proposed in this section are independent of a specific progress measure.

The y-axis is a metric for classification quality.
There are two ways to evaluate it for imbalanced class distributions: by computing a summary statistic on the binary confusion matrix, or by assessing the ranking induced by the decision function.

\subsection{One-Class Evaluation Metrics}

In this article, we use the Matthews Correlation Coefficient (MCC) and Cohen's kappa to evaluate the binary output. 
They can be computed from the confusion matrix.
MCC returns values in $[-1, +1]$, where high values indicate good classification on both classes, $0$ equals a random prediction, and $-1$ is the total disagreement between classifier and ground truth.
kappa returns $1$ for a perfect agreement with the ground truth and $0$ for one not better than a random allocation.

One can also use the distance to the decision boundary to rank observations.
The advantage is the finer differentiation between strong and less strong outliers.
A common metric is the area under the ROC curve (AUC) which has been used in other outlier-detection benchmarks \cite{Campos2016-ux}.
An interpretation of the AUC is the probability that an outlier is ranked higher than an inlier.
So an AUC of $1$ indicates a perfect ranking; $0.5$ means that the ranking is no better than random.

If the data set is large, users tend to only inspect the top of the ranked list of observations.
Then it can be useful to use the partial AUC (pAUC).
It evaluates classifier quality at thresholds on the ranking where the false-positive rate (FPR) is low.
An example for using pAUC to evaluate one-class active learning is \cite{Gornitz2013-ed}.

\subsection{Summary of the Active-Learning Curve}

The visual comparison of active learning via progress plots does not scale with the number of experiments.
For instance, our benchmark would require to compare \numexperiments different learning curves; this is prohibitive.
For large-scale comparisons, one should instead summarize a progress curve.
Recently, \emph{true performance of the selection strategy (TP)} has been proposed as a summary of increase and decrease of classifier performance over the number of iterations~\cite{Reyes2018-ey}.
However, TP is a single aggregate measure, which is likely to overgeneralize and is difficult to interpret.
For a more comprehensive evaluation, we therefore propose to use several summary statistics.
Each of them captures some characteristic of the learning progress and has a distinct interpretation.

We use $QM(k)$ for the quality metric $QM$ at the active learning progress $k$.
We use $\mathcal{L^{\text{init}}}$ and $\mathcal{L^{\text{end}}}$ to refer to the labeled examples at $t_{init}$ and $t_{end}$.\\

\emph{Start Quality (SQ):} The Start Quality is the baseline classification quality before the active learning starts, i.e., the quality of the base learner at the initial setup
\begin{equation*}
	SQ = QM(t_{init}).
\end{equation*}

\emph{Ramp-Up (RU):} The ramp-up is the quality increase after the initial $k$ progress steps.
A high RU indicates that the query strategy adapts well to the initial setup
\begin{equation*}
	RU(k) = QM(t_{k}) - QM(t_{init}).
\end{equation*}

\emph{Quality Range (QR):} The Quality Range is the increase in classification quality over an interval $[t_i, t_j]$.
A special case is $QR(\text{init}, \text{end})$, the overall improvement achieved with an active learning strategy
\begin{equation*}
	QR(i,j) = QM(t_{i}) - QM(t_{j}).
\end{equation*}

\emph{Average End Quality (AEQ):} In general, the progress curve is non-monotonic because each query introduces a selection bias in the training data.
So a query can lead to a quality decrease.
The choice of $t_{end}$ often is arbitrary and can coincide with a temporary bias.
So we propose to use the Average End Quality to summarize the classification quality for the final $k$ progress steps
\begin{equation*}
	AEQ(k) = \frac{1}{k} \sum_{i=1}^{k} QM(t_{end - k}).
\end{equation*}

\emph{Learning Stability (LS):} Learning Stability summarizes the influence of the last $k$ progress steps on the quality.
A high LS indicates that one can expect further improvement from continuing the active learning cycle.
A low LS on the other hand indicates that the classifier tends to be saturated, i.e., additional feedback does not increase the quality.
We define LS as the ratio of the average QR in the last $k$ steps over the average QR between init and end
\begin{equation*}
LS(k) = \begin{cases}
	\frac{QR(\text{end}-k, \text{end})}{k} \big/ \frac{QR(\text{init}, \text{end})}{\lvert \mathcal{L^{\text{end}}} \setminus \mathcal{L^{\text{init}}}\rvert}\ \text{if} \ QR(\text{init}, \text{end}) > 0 \\
	0\ \ \ \ \ \ \ \ \text{otherwise}.
\end{cases}
\end{equation*}

\emph{Ratio of Outlier Queries (ROQ):} The Ratio of Outlier Queries is the proportion of queries that the oracle labels as outlier
\begin{equation*}
ROQ = \frac{\lvert\mathcal{L_{\outlier}^{\text{end}}} \setminus \mathcal{L_{\outlier}^{\text{init}}} \rvert}{\lvert\mathcal{L}^{\text{\text{end}}} \setminus \mathcal{L}^{\text{init}} \rvert}.
\end{equation*}

\noindent
In practice, the usefulness of a summary statistic to select a good active learning strategy depends on the learning scenario.
For instance, ROQ is only meaningful if the user has a specific interest in observations from the minority class.\\

We conclude the discussion of summary statistics with two comments.
The first comment is on \emph{Area under the Learning Curve} (AULC), which also can be used to summarize active learning curves~\cite{Cawley2011-of,Reyes2018-ey}.
We deliberately choose to not include AULC as a summary statistic for the following reasons.
First, active learning is discrete, i.e., the minimum increment during learning is one feedback label.
But since the learning steps are discrete, the \enquote{area} under the curve is equivalent to the sum of the quality metric over the learning progress $\sum_{i=\text{init}}^{end} QM(t_{i})$.
In particular, approximating the AULC by, say, a trapezoidal approximation~\cite{Reyes2018-ey} is not necessary.
Second, AULC is difficult to interpret.
For instance, two curves can have different shapes and end qualities, but yet result in the same AULC value.
In our article we therefore rely on AEQ and SQ, which one can interpret as a partial AULC, with distinct interpretation.

Our second comment is using summary statistics to select different query strategies for different phases of the active learning cycle is conceivable in principle.
For instance, one could start the cycle with a good RU and then switch to a strategy with a good AEQ.
However, this leads to further questions, e.g., how to identify a good switch point, that go beyond this current article.

\subsection{Split Strategy}\label{sec:split-strategy}
A split strategy specifies how data is partitioned between training and testing.
With binary classifiers, one typically splits data into disjunct train and a test partition, which ideally are identically distributed.
However, since outliers do not come from a joint distribution, measuring classification quality on an independent test set is misleading.
In this case, one may measure classification quality as the resubstitution error, i.e., the classification quality on the training data.
This error is an optimistic estimate of classification quality.
But we deem this shortcoming acceptable if only a small percentage of the data has been labeled.

The learning objective should also influence how the data is split.
For instance, if the learning objective is to reliably estimate the majority-class distribution, one may restrict the training set to inliers (cf.\ \cite{Ghasemi2011-fl,Ghasemi2011-tv}).
Three split strategies are used in the literature.

\begin{enumerate}[leftmargin = 3.8\parindent, rightmargin = 2em, topsep = 1ex, partopsep=1ex, align=left, labelwidth=2.3em]
	\item[(Sh)] \underline{S}plit \underline{h}oldout: Model fitting and query selection on the training split, and testing on a distinct holdout sample.
	\item[(Sf)] \underline{S}plit \underline{f}ull: Model fitting, query selection and testing on the full data set.
	\item[(Si)] \underline{S}plit \underline{i}nlier: Like Sf, but model fitting on labeled inliers only.
\end{enumerate}

\begin{figure}[t!]
	\centering
	\includegraphics[width=0.65\linewidth]{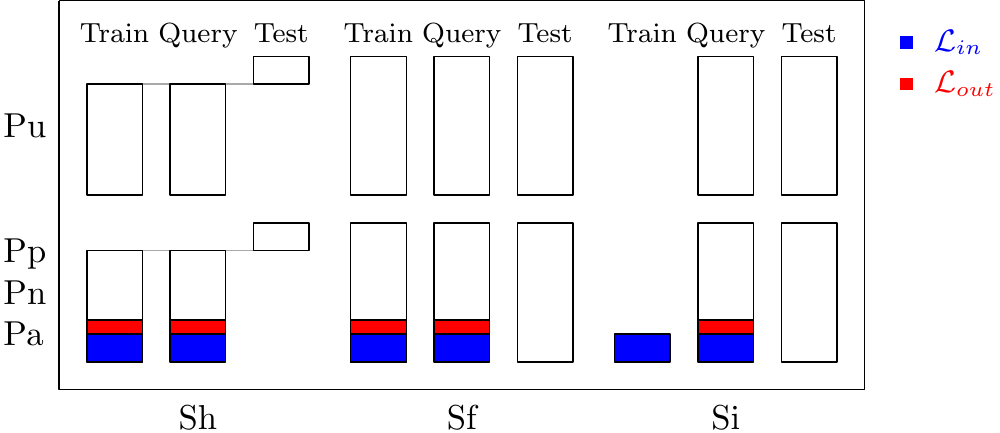}
	\caption{Illustration of initial pool and split strategies. The blue and the red proportion indicate the labeled inliers and outliers in the initial pools that make up for p-\% of the full data~(Pp), a fixed number of observations~(Pn) or the number of attributes~(Pa).}\label{fig:learning_scenarios}
\end{figure}

\noindent
Split strategies increase the complexity of evaluating active learning, since they must be combined with an initial pool strategy.
Most combinations of split strategies and initial pool strategies are conceivable.
Only no labels (Pu) does not work with a split strategy that fits a model solely on inliers (Si) -- the train partition would be empty in this case.
\autoref{fig:learning_scenarios} is an overview of all combinations of an initial pool strategy and a split strategy.

\subsection{Limitations}
\label{sec:building-block-combinations}

Initial setups, split strategies, base learners and query strategies (QS) all come with prerequisites.
One cannot combine them arbitrarily, because some of the prerequisites are mutually exclusive, as follows.
\begin{itemize}
	\item Pu rules out any data-based QS. 
	This is because data-based QS require labeled observations for the density estimations.
	\item Kernel-density estimation requires the number of labeled observations to be at least equal to the number of attributes.
	A special case is $\tau_{\text{ML}}$, Which requires $|\mathcal{L}_{outlier}| \geq M$.
	As a remedy, one can omit the subtrahend in \autoref{eq:tau-cv-in} and \autoref{eq:tau-cv-out} in this case.
	\item Fully unsupervised base learners, e.g., SVDD, are only useful when the learning objective is a robust estimate of the majority distribution, and when the split strategy is Si.
	The reason is that feedback can only affect the classifier indirectly, by changing the composition of the training data $\mathcal{L}_\inlier$.
	\item A combination of Pu and Si is not feasible, see~\autoref{sec:split-strategy}.
\end{itemize}
\autoref{tab:tau_pool} is an overview of the feasibility of query strategies.
In what follows, we only consider feasible combinations.

\begin{table*}
	\centering
	\caption{Overview over the number of labels required by different query strategies; $M$ is the number of attributes.\newline
		Feasible: $\checkmark$, feasible with modification: $(\checkmark)$, not feasible: $\times$.}
	\label{tab:tau_pool}
	\resizebox{\columnwidth}{!}{
		\begin{tabular}{lcccccccccc} \toprule
			Scenario & $\tau_{\text{MM}}$ &  $\tau_{\text{EME}}$ & $\tau_{\text{EME}}$ & $\tau_{\text{ML}}$ & $\tau_{\text{HC}}$ & $\tau_{\text{DB}}$ & $\tau_{\text{NB}}$ & $\tau_{\text{BNC}}$ & $\tau_{\text{rand}}$ & $\tau_{\text{rand-out}}$ \\ \midrule
			$|\mathcal{L}_{in}| = 0 \,\wedge\, |\mathcal{L}_{out}| = 0$ & $\times$ & $\times$ & $\times$ & $\times$ & $\times$ & $\checkmark$ & $\checkmark$ & $\checkmark$ & $\checkmark$ & $\checkmark$ \\\midrule
			$|\mathcal{L}_{in}| \geq M \,\wedge\, |\mathcal{L}_{out}| = 0$ & $\checkmark$ & $\checkmark$ & $\checkmark$ & $(\checkmark)$ & $\checkmark$ & $\checkmark$ & $\checkmark$ & $\checkmark$ & $\checkmark$ & $\checkmark$ \\\midrule
			$|\mathcal{L}_{in}| \geq M \,\wedge\, |\mathcal{L}_{out}| \geq M$ & $\checkmark$ & $\checkmark$ & $\checkmark$ & $\checkmark$ & $\checkmark$ & $\checkmark$ & $\checkmark$ & $\checkmark$ & $\checkmark$ & $\checkmark$ \\\bottomrule
		\end{tabular}
	}
\end{table*}

\section{Benchmark}
\label{sec:experiments-and-results}

The plethora of ways to design and to evaluate AL systems makes selecting a good configuration for a specific application difficult.
Although certain combinations are infeasible, the remaining options are still too numerous to analyze.
This section addresses question \emph{Comparison} and provides some guidance how to navigate the overwhelming design space.
We have implemented the base learners, the query strategies and the benchmark setup in \emph{Julia}~\cite{bezanson2017julia}.
Our implementation, the raw results of all settings and notebooks to reproduce experiments and evaluation are publicly available at \website.

We begin by explaining our experiments conducted on well-established benchmark data sets for outlier detection~\cite{Campos2016-ux}.
In total, we run experiments on over \numexperiments configurations:
72,000 configurations in \autoref{sec:sub:evaluation-metric} to \autoref{sec:sub:initial-pool-strategies} are the cross product of 20 data sets, 3 resampled versions, 3 split strategies, 4 initial pool strategies, 5 models with different parametrization, 2 kernel parameter initializations and 10 query strategies;
12,000 additional configurations in \autoref{sec:sub:query-strategy} are the cross product of 20 data sets with 3 resampled versions each, 2 models, 2 kernel parameter initializations, 5 initial pool resamples and 10 query strategies.
\autoref{tab:experimental-setup} lists the experimental space, see~\autoref{app:experimental_setup} for details.

Each specific experiment corresponds to a practical decision which query strategy to choose in a specific setting.
We illustrate this with the following example. 

\newcommand{\qsa}{$\tau_{\text{HC}}$\xspace}
\newcommand{\qsb}{$\tau_{\text{NB}}$\xspace}

\begin{example}[Experiment Run]
	Assume the data set is Arrhythmia, and there are no initial labels, i.e., the initial pool strategy is Pu, and data-based QS are not applicable.
	The classifier is SVDDneg, and we use Sf to evaluate the classification quality.
	Our decision is to choose \qsa and \qsb as potential query strategies and to terminate the active-learning cycle after 100 iterations.
	
	\begin{figure}
		\centering
		\includegraphics[width=0.7\linewidth]{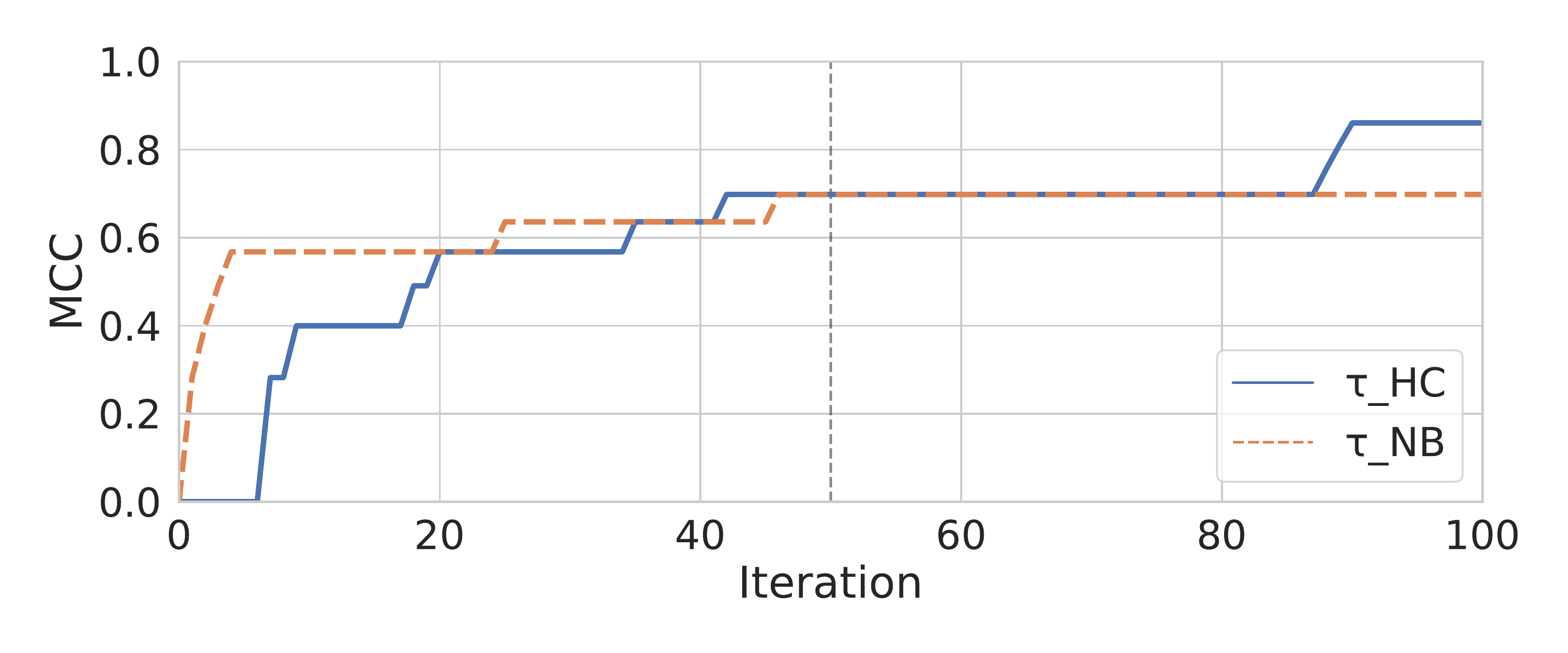}
		\caption{Comparison of the progress curves for two query strategies on Arrhythmia.}
		\label{fig:eval_progress_curve_example}
	\end{figure}
	
	\autoref{fig:eval_progress_curve_example} graphs the progress curves for both query strategies.
	A first observation is that it depends on the progress which one is better.
	For example, \qsb results in a better MCC after 10 iterations, while \qsb is superior after 90 iterations.
	After 50 iterations, both \qsa and \qsb perform equally well.
	Until iteration 60, the learning stability (LS) decreases to 0, which speaks for stopping.
	Indeed, although there is some increase after 60 iterations, it is small compared to the overall improvement.
	
	For a more differentiated comparison, we now look at several progress curve summaries.
	If only the final classification quality is relevant, i.e., the budget is fixed to 100 observations, \qsa is preferred because of higher EQ and AEQ values.
	For a fast adaption, one should prefer \qsb with RU(5) = 0.57, compared to a RU(5) = 0.00 for \qsa.
	Regarding the outlier ratio, both query strategies perform similarly with \SI{7}{\percent} and \SI{9}{\percent}.
	Users can now weigh these criteria based on their preferences to decide on the most appropriate query strategy.
\end{example}

\begin{table}
	\caption{Overview on experimental setup.}\label{tab:experimental-setup}
	\centering
	\begin{tabular}{L{0.2\columnwidth} L{0.7\columnwidth}}
		\toprule
		Dimension & Configuration \\
		\midrule 
		Initial Pools & Pu, Pp~($p = 0.1$), Pn~($n=25$), Pa\\\midrule
		Split Strategy & Sf, Sh (80\%~train, 20\%~test), Si\\\midrule
		Base Learner & SVDD, SVDDneg, SSAD~($\kappa = 1.0, 0.5, 0.1$)\\\midrule
		Kernel Initialization & Wang, Scott\\\midrule
		Query strategy & $\tau_\text{MM}$, $\tau_\text{EMM}$, $\tau_\text{EME}$, $\tau_\text{ML}$, $\tau_\text{HC}$, $\tau_\text{DB}$, $\tau_\text{NB}$, $\tau_\text{BNC}$, $\tau_\text{rand}$, $\tau_\text{rand-out}$ \\
		\bottomrule 
	\end{tabular}
\end{table}

\noindent
In our benchmark, we strive for general insights and trends regarding such decisions.
In the following, we first discuss assumptions we make for our benchmark and the experiment setup.
We then report on results.
Finally, we propose guidelines for outlier detection with active learning and discuss extensions to our benchmark towards conclusive decision rules.

\subsection{Assumptions}

We now specify the assumptions behind our benchmark.

\emph{General Assumptions.}
In our benchmark, we focus on \enquote{sequential class label} as the \emph{feedback type}.
We set the \emph{feedback budget} to a fixed number of labels a user can provide.
The reason for a fixed budget is that the number of queries in an active learning cycle depends on the application, and estimating active learning performance at runtime is difficult~\cite{Kottke2019-ua}.
There is no general rule how to select the number of queries for evaluation.
In our experiments, we perform 50 iterations.
Since we benchmark on many publicly available data sets, we do not have any requirements regarding interpretability. 
Instead, we rely on the ground truth shipped with the data sets to simulate a perfect oracle.

\emph{Specific Assumptions.}
We have referred to specific assumptions throughout this article and explained how they affect the building blocks and the evaluation. 
For clarity, we briefly summarize them.
For the \emph{class distribution}, we assume that outliers do not have a joint distribution.
The primary learning objective is to improve the accuracy of the classifier.
However, we also use the \emph{ROQ} summary statistic to evaluate whether a method yields a high proportion of queries from the minority class.
For the initial setup, we do not make any further assumptions.
Instead, we compare the methods on all feasible combinations of initial pools and split strategies.\\

\subsection{Experimental Setup}
\label{app:experimental_setup}

Our experiments cover several instantiations of the building blocks.
\autoref{tab:datasets} lists the data sets, and \autoref{tab:experimental-setup} lists the experimental space.
For each data set we use three resampled versions with an outlier percentage of 5\% that have been normalized and cleaned from duplicates.
We have downsampled large data sets to $N = 1000$.
This is comparable to the size of the data sets used in previous work for active learning for one-class classification.
Additionally, one may use sampling techniques for one-class classifiers to scale to large data sets, e.g., \cite{Krawczyk2018-sl,Li2011-mb}.
However, further studying the influence of the data set size on the query strategies is out of the scope of this article.

\begin{table}[ht]
	\centering
	\caption{Overview on data sets.}\label{tab:datasets}
	\begin{tabular}{lrr}
		\toprule
		Dataset & Observations (N) & Attributes (M) \\
		\midrule
			ALOI &        1000 &              27 \\
			Annthyroid &        1000 &              21 \\
			Arrhythmia &         240 &             259 \\
			Cardiotocography &        1000 &              21 \\
			Glass &         180 &               7 \\
			HeartDisease &         140 &              13 \\
			Hepatitis &          60 &              19 \\
			Ionosphere &         237 &              32 \\
			KDDCup99 &        1000 &              40 \\
			Lymphography &         120 &               3 \\
			PageBlocks &        1000 &              10 \\
			PenDigits &         400 &              16 \\
			Pima &         520 &               8 \\
			Shuttle &         260 &               9 \\
			SpamBase &        1000 &              57 \\
			Stamps &         320 &               9 \\
			WBC &         200 &               9 \\
			WDBC &         200 &              30 \\
			WPBC &         159 &              33 \\
			Waveform &        1000 &              21 \\
		\bottomrule
	\end{tabular}
\end{table}

\emph{Parameters:} Parameter selection for base learners and query strategies is difficult in an unsupervised scenario.
One must rely on heuristics to select the kernel and cost parameters for the base-learners, see \autoref{sec:base-learner}.
We use Scott's rule of thumb~\cite{scott2015multivariate} and state-of-the-art self-adapting data shifting by Wang et al.\ \cite{wang2018hyperparameter} for the kernel parameter $\gamma$.
For cost $C$ we use the initialization strategy of Tax et al.\ \cite{Tax2004-ss}.
For SSAD, the authors suggest to set the trade-off parameter $\kappa = 1$~\cite{Gornitz2013-ed}.
However, preliminary experiments of ours indicate that SSAD performs better with smaller parameter values in many settings.
Thus, we include $\kappa = 0.1$ and $\kappa = 0.5$ as well.
For the query strategies, the selection of strategy-specific parameters is described in \autoref{sec:query-strategies}.
The data-based query strategies use the same $\gamma$ value for kernel density estimation as the base learner.

\subsection{Results}

We now discuss general insights and trends we have distilled from the experiments.
We start with a broad overview and then fix some experimental dimensions step by step to analyze specific regions of the experimental space.
We begin by comparing the expressiveness of evaluation metrics and the influence of base learner parametrization.
Then we study the influence of the split strategy, the initial pool strategy, and the query strategy on result quality.

\subsubsection{Evaluation Metric}\label{sec:sub:evaluation-metric}

\begin{figure}[t!]
	\centering
	\includegraphics[width=0.7\linewidth]{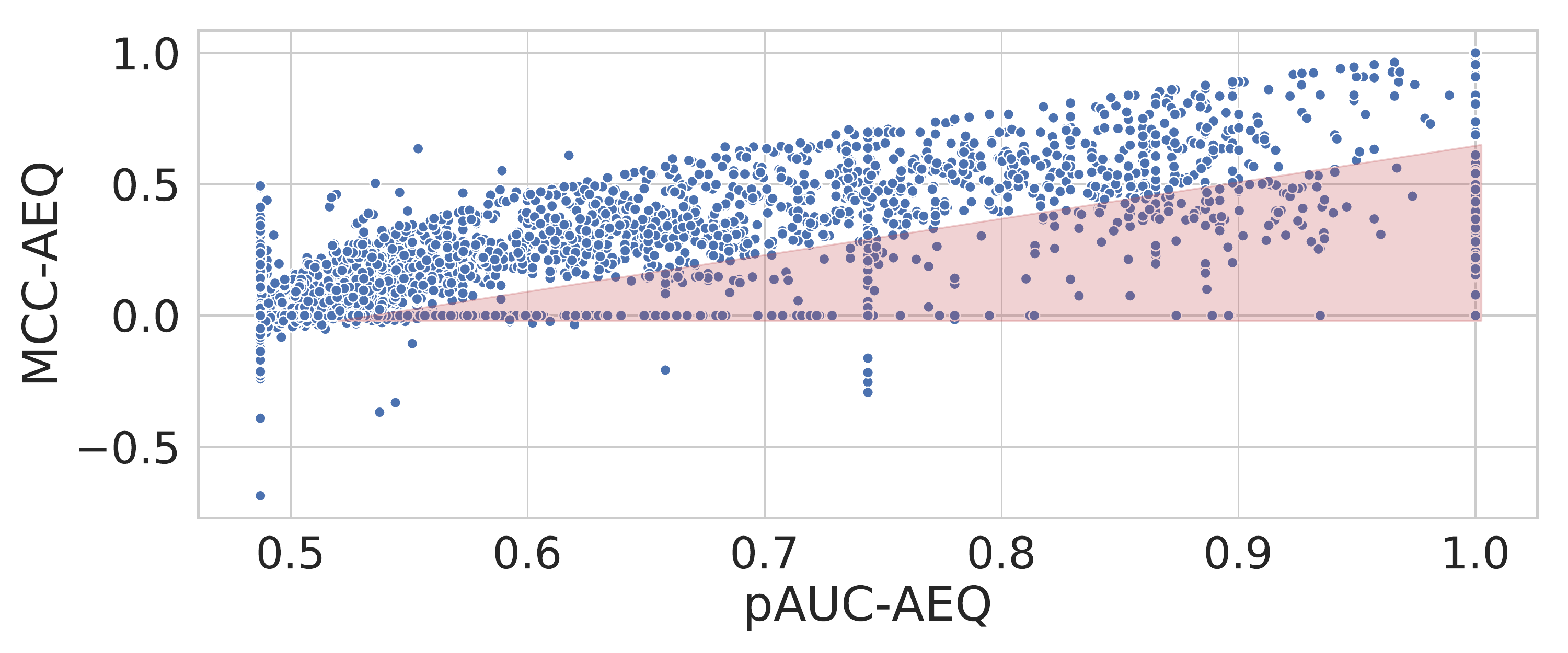}
	\caption{Comparison of pAUC and MCC. Each point corresponds to an experimental run.}
	\label{fig:eval_metric}
\end{figure}
\begin{table}
	\centering
	\caption{Pearson correlation of AEQ (k=5) for different evaluation metrics.}\label{tab:em_correlation}
	\begin{tabular}{lrrrr}
		\toprule
		{} &       MCC &     kappa &       AUC &      pAUC \\
		\midrule
		MCC   &  1.00 &   0.98 &  0.63 &  0.78 \\
		kappa &  0.98 &   1.00 &  0.59 &  0.76 \\
		AUC   &  0.63 &   0.59 &  1.00 &  0.73 \\
		pAUC  &  0.78 &   0.76 &  0.73 &  1.00 \\
		\bottomrule
	\end{tabular}
\end{table}
Recall that our evaluation metrics are of two different types: ranking metrics (AUC and pAUC) and metrics based on the confusion matrix (kappa, MCC).
On all settings, metrics of the same type have a high correlation for AEQ, see~\autoref{tab:em_correlation}.
So we simplify the evaluation by selecting one metric of each type.

Further, there is an important difference between both types. 
\autoref{fig:eval_metric} depicts the AEQ for pAUC and MCC.
For high MCC values, pAUC is high as well.
However, high pAUC values often do not coincide with high MCC values, please see the shaded part of the plot.
In the extreme cases, there even are instances where pAUC = 1 and MCC is close to zero.
In this case, the decision function induces a good ranking of the observations, but the actual decision boundary does not discern well between inliers and outliers.
An intuitive explanation is that outliers tend to be farthest from the center of the hypersphere.
Because pAUC only considers the top of the ranking, it merely requires a well-located center to arrive at a high classification quality.
But the classifier actually may not have fit a good decision boundary.

Our conclusion is that pAUC and AUC may be misleading when evaluating one-class classification.
Hence, we only use MCC from now on.

\begin{figure}
	\centering
	\includegraphics[width=0.7\linewidth]{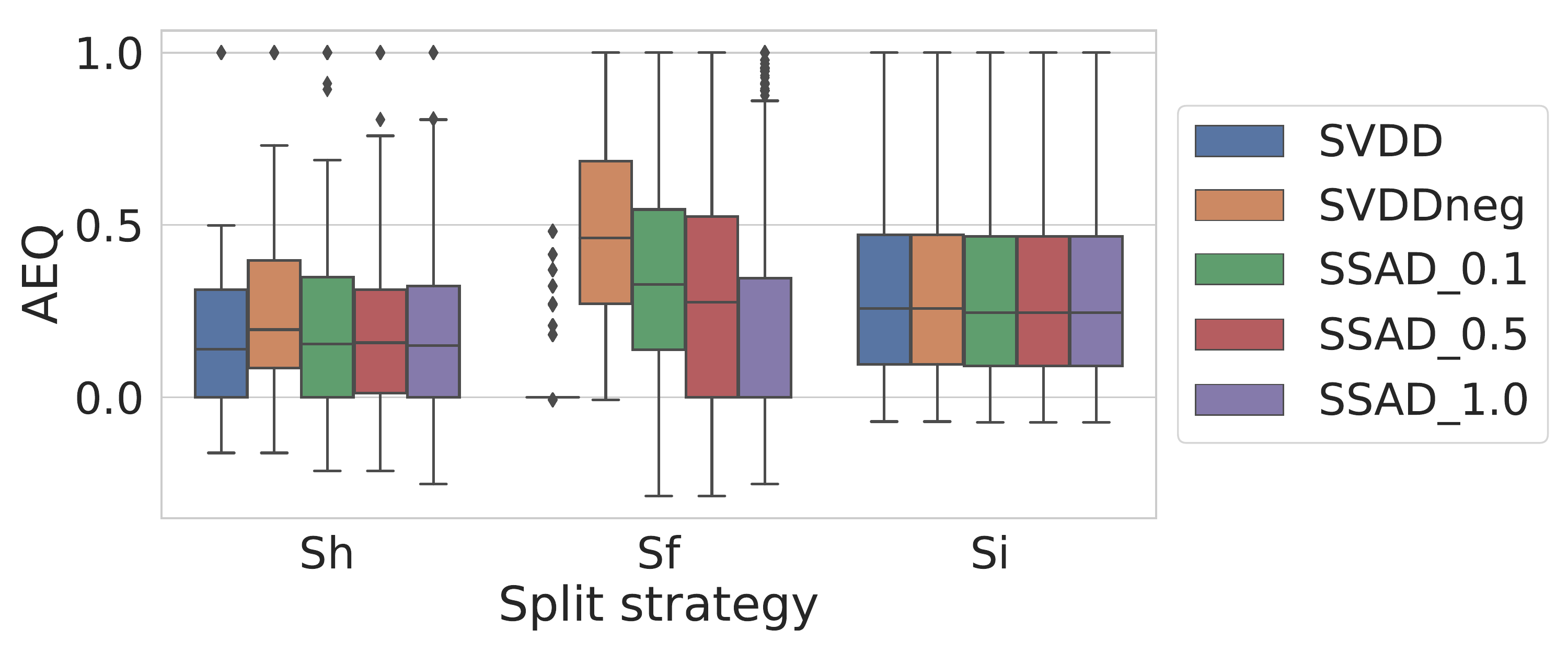}
	\caption{Evaluation of the AEQ (k=5) for different split strategies grouped by base learner.}
	\label{fig:eval_split_strategies}
\end{figure}

\subsubsection{Influence of Kernel Parameter}

\begin{table}[t]
	\centering
	\caption{Median AEQ (k=5) and SQ for different gamma initialization strategies.}\label{tab:gamma}
	\begin{tabular}{lrrrr}
		\toprule
		Model & \multicolumn{2}{c}{SQ} & \multicolumn{2}{c}{AEQ} \\
		& Scott & WangTax & Scott & WangTax \\
		\midrule
		SVDD &  0.06 &    0.06 &  0.09 &    0.10 \\
		SVDDneg &  0.09 &    0.14 &  0.21 &    0.31 \\
		SSAD\_0.1 &  0.07 &    0.11 &  0.15 &    0.24 \\
		SSAD\_0.5 &  0.06 &    0.06 &  0.12 &    0.22 \\
		SSAD\_1.0 &  0.04 &    0.08 &  0.10 &    0.15 \\
		\bottomrule
	\end{tabular}
\end{table}

Recall that the kernel parameter influences the flexibility of the decision boundary; high values correspond to more flexible boundaries.
Our hypothesis is that a certain flexibility is necessary for models to adapt to feedback.

\autoref{tab:gamma} shows the SQ and AEQ for two heuristics to initialize $\gamma$.
In both summary statistics, Wang strategy outperforms the more simple Scott rule of thumb significantly on the median over all data sets and models.
A more detailed analysis shows that there are some data sets where Scott outperforms Wang, e.g., KDD-Cup.
However, there are many instances where Wang performs well, but Scott results in very poor active learning quality.
For instance, the AEQ on Glass for Scott is $0.06$, and for Wang $0.45$.
We hypothesize that this is because Scott yields very low $\gamma$ values for all data sets, and the decision boundary is not flexible enough to adapt to feedback.
The average value is $\gamma=0.77$ for Scott and $\gamma=5.90$ for Wang.

We draw two conclusions from these observations.
First, the choice of $\gamma$ influences the success of active learning significantly.
When the value is selected poorly, active learning only results in minor improvements on classification quality -- regardless of the query strategy.
Second, Wang tends to select better $\gamma$ values than Scott, and we use it as the default in our experiments.
Our observations also motivate further research on how to select the parameters in an active learning setting.
However, studying this issue goes beyond the scope of our article.

\subsubsection{Split Strategies}

Our experiments show that split strategies have a significant influence on classification quality.
\autoref{fig:eval_split_strategies} graphs the AEQ for the different split strategies grouped by base learners.

We first compare the three split strategies.
For Sh, the AEQ on the holdout sample is rather low for all base learners. 
For Sf, SVDDneg and SSAD\_0.1 achieve high quality.
Some of this difference may be explained by the more optimistic resubstitution error in Sf.
However, the much lower AEQ in Sh, for instance for SVDDneg, rather confirms that outliers do not follow a homogeneous distribution (cf.\ \autoref{sec:learning-scenario}).
In this case, the quality on the holdout sample is misleading.

For Si, all classifiers yield about the same quality.
This is not surprising.
The classifiers are trained on labeled inliers only. 
So the optimization problems for the base learners coincide.
The average quality is lower than with Sf, because the training split only contains a small fraction of the inliers.
Based on all this, we question whether Si leads to an insightful evaluation, and we exclude Si from now on.

Next, we compare the quality of the base learners.
For Sf, SVDD fails because it is fully unsupervised, i.e., cannot benefit from feedback.
For SSAD, the quality fluctuates with increasing $\kappa$.
Finding an explanation for this is difficult.
We hypothesize that this is because SSAD overfits to the feedback for high $\kappa$ values.
For Sf, $\kappa = 0.1$ empirically is the best choice.

In summary, the split strategy has a significant effect on classification quality.
SVDDneg and SSAD\_0.1 for Sf yield the most reasonable results.
We fix these combinations for the remainder of this section.\\

\begin{table}[t!]
	\centering
	\caption{Comparison of SQ and AEQ (k=5) for different initial pool strategies, Pp = \SI{10}{\percent}, Pn = 20 observations.}
	\label{tbl:eval_initial_pool_strategies}
	\begin{tabular}{lC{1cm}cC{1.3cm}cc}
		\toprule
		Data set & Initial Pool &  n &  Initially labeled & SQ & AEQ \\
		\midrule
		ALOI &            Pn &          1000 &                 20 &  0.00 &  0.14 \\
		&            Pp &          1000 &                100 &  0.17 &  0.22 \\
		WBC &            Pn &           200 &                 20 &  0.31 &  0.74 \\
		&            Pp &           200 &                 20 &  0.31 &  0.72 \\
		\bottomrule
	\end{tabular}
\end{table}

\subsubsection{Initial Pool Strategies}\label{sec:sub:initial-pool-strategies}

The initial pool strategy specifies the number of labeled observations at $t_{init}$.
Intuitively, increasing it should increase the start quality, as more information on the data is available to the classifier.
If the initial pool is representative of the underlying distribution, little benefit can be expected from active learning.

\begin{figure}[t!]
	\centering
	\includegraphics[width=0.7\linewidth]{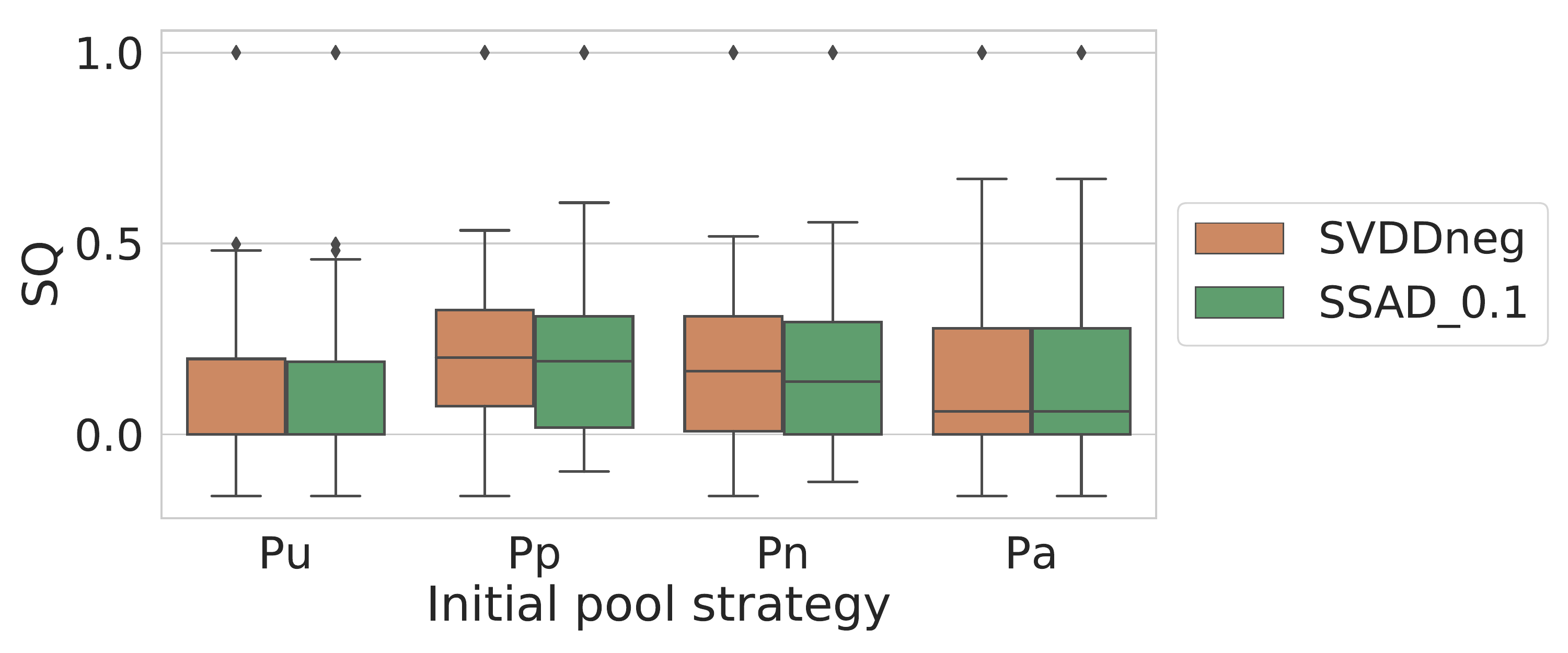}
	\caption{Evaluation of the different initial pool strategies.}
	\label{fig:eval_initial_pool_strategies}
\end{figure}

Our results confirm this intuition.
\autoref{fig:eval_initial_pool_strategies} shows the SQ for the initial pool strategies grouped by SVDDneg and SSAD\_0.1.
For Pu, there are no labeled observations, and the corresponding SQ is low.
When labeled data is available, Pp tends to yield a better SQ than Pn.
However, the figure is misleading, because the actual number of labels depends on the data set.
This becomes clear when looking at ALOI and WBC, see \autoref{tbl:eval_initial_pool_strategies}.
For WBC, Pp and Pn result in a similar number of initial labels.
For ALOI however, the number of labels with Pp is five times larger than with Pn.
So the SQ on ALOI is higher for Pp, but AEQ is only slightly higher than SQ.
This means that active learning has comparatively little effect.
Pa has a technical motivation, i.e., it is the minimal number of labels required by the data-based strategies.
This strategy is not feasible for data sets where the number of attributes is larger than the number of observations.
Other than this, the interpretation of Pa is similar to Pp with $p = \frac{M}{N}$.

In summary, different initial pool strategies lead to substantially different results.
We deem Pn more intuitive than Pp when reporting results, since the size of the initial sample, and hence the initial labeling effort is explicit.
In any case, one must carefully state how the initial sample is obtained.
Otherwise, it is unclear whether high quality according to AEQ is due to the query strategy or to the initial pool.

\subsubsection{Query Strategy}\label{sec:sub:query-strategy}
We have arrived at a subset of the experimental space where comparing different query strategies is reasonable.
To do so, we fix the initial pool strategy to Pn with $n=25$.
In this way, we can include the data-based QS which all require initial labels.
We obtain the initial pool by uniform stratified sampling.
Additionally, we exclude the Hepatitis data set because it only contains 60 observations; this is incompatible with 20 initially labeled observations and 50 iterations.
We repeat each setting 5~times and average the results to reduce the bias of the initial sample.

\begin{table*}
	\centering
	\caption{Median QR on Pn for different query strategies over 30 settings; highest values per data set in bold.}
	\label{tbl:eval_query_strategies}
	\resizebox{\columnwidth}{!}{
		\begin{tabular}{l|cccc|cc|cc|cc}
			\toprule
			\multicolumn{1}{c}{} & \multicolumn{4}{c}{data-based} & \multicolumn{2}{c}{model-based} & \multicolumn{2}{c}{hybrid} & \multicolumn{2}{c}{baselines} \\
			Data set &  $\tau_\text{MM}$ &  $\tau_\text{EMM}$ &  $\tau_\text{EME}$ &  $\tau_\text{ML}$ &  $\tau_\text{HC}$ &  $\tau_\text{DB}$ &  $\tau_\text{NB}$ &  $\tau_\text{BNC}$ &  $\tau_\text{rand}$ &  $\tau_\text{rand-out}$ \\
			\midrule
             ALOI &                 - &                         - &                          - &               - &               \textbf{0.30} &                 \textbf{0.30} &                  0.28 &                            0.00 &       0.00 &              0.00 \\
Annthyroid &                 - &                         - &                          - &               - &               \textbf{0.28} &                 0.20 &                  \textbf{0.28} &                            0.24 &       0.09 &              0.14 \\
Arrhythmia &                 - &                         - &                          - &               - &               \textbf{0.73} &                 \textbf{0.73} &                  0.53 &                            0.40 &       0.00 &              0.70 \\
Cardiotocography &                 - &                         - &                          - &               - &               0.42 &                 \textbf{0.46} &                  \textbf{0.46} &                            0.05 &       0.08 &              0.10 \\
Glass &               0.0 &                      0.56 &                       0.14 &            0.67 &               \textbf{0.67} &                 \textbf{0.67} &                  0.55 &                            0.55 &       0.24 &              0.24 \\
HeartDisease &               0.0 &                      0.00 &                       0.28 &            0.34 &               0.23 &                 0.23 &                  0.32 &                            0.29 &       0.16 &              \textbf{0.36} \\
Ionosphere &                 - &                         - &                          - &               - &               0.65 &                 \textbf{0.72} &                  0.53 &                            0.12 &       0.21 &              0.35 \\
KDDCup99 &                 - &                         - &                          - &               - &               0.36 &                 \textbf{0.53} &                  \textbf{0.53} &                            0.33 &       0.14 &              0.14 \\
Lymphography &               0.0 &                      0.00 &                       0.51 &            \textbf{0.67} &               0.40 &                 0.41 &                  0.41 &                            0.43 &       0.41 &              0.34 \\
PageBlocks &               0.0 &                      0.00 &                       0.01 &            0.00 &               0.09 &                 0.22 &                  \textbf{0.24} &                            0.11 &       0.01 &              0.08 \\
PenDigits &               0.0 &                      0.00 &                       0.00 &            0.22 &               0.19 &                 0.19 &                  0.50 &                            0.09 &       0.16 &              \textbf{0.78} \\
Pima &               0.0 &                      0.00 &                       0.00 &            0.39 &               0.28 &                 0.35 &                  0.35 &                            0.19 &       0.19 &              \textbf{0.50} \\
Shuttle &               0.0 &                      0.00 &                       0.38 &            0.37 &               0.57 &                 0.57 &                  0.58 &                            0.61 &       0.27 &              \textbf{0.75} \\
SpamBase &                 - &                         - &                          - &               - &               0.28 &                 0.28 &                  \textbf{0.29} &                            0.10 &       0.00 &              0.17 \\
Stamps &               0.0 &                      0.00 &                       0.24 &            0.62 &               0.61 &                 \textbf{0.65} &                  0.56 &                            0.47 &       0.24 &              0.00 \\
WBC &               0.0 &                      0.00 &                       0.58 &            \textbf{0.69} &               \textbf{0.69} &                 \textbf{0.69} &                  0.39 &                            0.13 &       0.31 &              \textbf{0.69} \\
WDBC &                 - &                         - &                          - &               - &               0.58 &                 0.58 &                  0.58 &                            0.23 &       0.39 &              \textbf{0.64} \\
WPBC &                 - &                         - &                          - &               - &               \textbf{0.26} &                 \textbf{0.26} &                  \textbf{0.26} &                            0.20 &       0.00 &              0.14 \\
Waveform &               0.0 &                      0.00 &                       0.14 &            0.00 &               \textbf{0.28} &                 \textbf{0.28} &                  \textbf{0.28} &                            0.21 &       0.06 &              0.00 \\
			\bottomrule
		\end{tabular}
	}
\end{table*}

\autoref{tbl:eval_query_strategies} shows the median QR(init, end) grouped by data set.
By design of the experiment, SQ is equal for all query strategies.
This means that AEQ coincides with QR.
On some data sets (indicated by \enquote{ - }), data-based query strategies fail.
The reason is that the rang of the matrix of observations, on which the kernel density is estimated, is smaller than $M$.
For the remaining data sets, we make two observations.
First, the QR achieved differs between data sets.
Some data sets, e.g., Annthyroid and PageBlocks, seem to be more difficult and only result in a small QR.
Second, the quality of a specific QS differs significantly between data sets.
For instance, $\tau_{\text{ML}}$ is the best strategy on Lymphography, but does not increase the classification quality on PageBlocks.
In several cases, $\tau_{\text{rand-out}}$ clearly outperforms the remaining strategies.
There neither is a QS category nor a single QS that is superior on all data sets.
This also holds for other metrics like RU and ROQ.

Next, runtimes for $\tau_{\text{ML}}$ are an order of magnitude larger than for all other strategies.
For PageBlocks, the average runtime per query selection for $\tau_{\text{ML}}$ is \SI{112}{\second}, compared to \SI{0.5}{\second} for $\tau_{\text{NB}}$.

To summarize, there is no one-fits-all query strategy for one-class AL.
The requirements for data-based query strategies may be difficult to meet in practice.
If the requirements are met, all model-based and hybrid strategies we have evaluated except for $\tau_{\text{BNC}}$ may be a good choice.
In particular, $\tau_{\text{DB}}$ and $\tau_{\text{rand-out}}$ are a good choice in the majority of cases.
They result in significant increases over 50 iterations for most data sets and scale well with the number of observations.
Even in the few cases where other query strategies outperform them, they still yield acceptable results.

\subsection{Guidelines and Decision Rules}\label{sec:towards-decision-rules}

The results from previous sections are conclusive and give way to general recommendations for outlier detection with active learning.
We summarize them as guidelines for the selection of query strategies and for the evaluation of one-class active learning.

\subsubsection{Guidelines}

\begin{enumerate}[label=(\roman*)]
	\item Learning scenario: We recommend to specify general and specific assumptions on the feedback process and the application.
	This narrows down the design space of building-block combinations.
	Regarding research, it may also help others to assess novel contributions more easily.
	\item Initial Pool: The initial pool strategy should either be Pu, i.e., a cold start without labels, or Pn with an absolute number of labels.
	It is important to make explicit if and how an initial sample has been obtained.
	\item Base Learner:
	A good parametrization of the base learner is crucial.
	To this end, selecting the bandwidth of the Gaussian kernel by self-adaptive data shifting~\cite{wang2018hyperparameter} works well.
	When parameters are well-chosen, SVDDneg is a good choice across data sets and query strategies.
	\item Query Strategies: Good choices across data sets are $\tau_{\text{DB}}$ and $\tau_{\text{rand-out}}$. 
	One should give serious consideration to random baselines, as they are easy to implement and outperform the more complex strategies in many cases.
	\item Evaluation:
	Progress curve summaries yield a versatile and differentiated view on the performance of active learning.
 	We recommend to use them to select query strategies for a specific use case.
	As the quality metric, we suggest to use MCC or kappa.
	Calculating this metric as a resubstitution error based on a Sf split is reasonable for outlier detection.
\end{enumerate}

\subsubsection{Beyond Guidelines}

From the results presented so far, one may also think about deriving a formal and strict set of rules to select an active learning method that are even more rigorous than the guidelines presented.
However, this entails major difficulties, as we now explain.
Addressing them requires further research that go beyond the scope of a comparative study.

\begin{enumerate}[label=(\roman*)]
	\item One can complement the benchmark with additional real-world data sets.
	But they are only useful to validate whether rules that have already been identified are applicable to other data as well.
	So, given our current level of understanding, we expect additional real-world data sets to only confirm our conclusion that formal rules currently are beyond reach.
	\item One may strive for narrow rules, e.g., rules that only apply to data with certain characteristics.
	This would require a different kind of experimental study, for instance with synthetic data.
	This also is difficult, for at least two reasons.
	First, it is unclear what interesting data characteristics would be in this case.
	Even if one can come up with such characteristics, it still is difficult to generate synthetic data with all these interesting characteristics.
	Second, reliable statements on selection rules would require a full factorial design of these characteristics.
	This entails a huge number of combinations with experiment runtimes that are likely to be prohibitive.
	To illustrate, even just 5 characteristics with 3 manifestations each result in a $3^5=243$  data sets instead of 20 data sets, and a total of 874,800 experiments -- an order of magnitude larger than the experiments presented here.
	Yet our experiments already have a sequential run time of around 482 days.
	\item One could strive for theoretical guarantees on query strategies.
	But the strategies discussed in \autoref{sec:sub:query-strategy} are heuristics and do not come with any guarantees.
	A discussion of the theoretical foundations of active learning may provide further insights.
	However, this goes beyond the scope of this current article as well.
\end{enumerate}
To conclude, deriving a set of formal rules based on our results is not within reach.
So one should still select active learning methods for a use case individually.
Our systematic approach from the previous sections does facilitate such a use-case specific selection.
It requires to carefully define the learning scenario and to use summary statistics for comparisons.

\section{Conclusions}

Active learning for outlier detection with one-class classifiers relies on several building blocks: the learning scenario, a base learner, and a query strategy.
While the literature features several approaches for each of the building blocks, finding a suitable combination for a particular use case is challenging.
In this article, we have approached this challenge, in two steps.
First, we provide a categorization of active learning for one-class classification and propose methods to evaluate active learning beyond progress curves.
Second, we have evaluated existing methods, using an extensive benchmark.
Our experimental results show that there is no one-fits-all strategy for one-class active learning.
Thus, we have distilled guidelines on how to select a suitable active learning method with specific use cases.
Our categorization, evaluation standards and guidelines give way to a more reliable and comparable assessment of active learning for outlier detection with one-class classifiers.
\begin{acks}

This work was supported by the German Research Foundation (DFG) as part of the Research Training Group GRK 2153: \textit{Energy Status Data -- Informatics Methods for its Collection, Analysis and Exploitation}.

\end{acks}

\bibliography{bib/ocal}
\bibliographystyle{ACM-Reference-Format}

\end{document}